\documentclass{article}
\usepackage{amsmath}
\usepackage{amssymb}
\usepackage{mathtools}
\usepackage{enumitem}
\usepackage{pdflscape} 
\usepackage{longtable}
\usepackage{graphicx}
\usepackage[a4paper]{geometry}
\usepackage[utf8]{inputenc}
\usepackage{array}
\usepackage{rotating}
\usepackage{xcolor,colortbl}
\usepackage{natbib}
\usepackage{hyperref}
\usepackage[font=small]{caption}
\usepackage{booktabs}
\usepackage{adjustbox}

\newcolumntype{P}[1]{>{\centering\arraybackslash}p{#1}}
\newcolumntype{M}[1]{>{\centering\arraybackslash}m{#1}}

\DeclareMathOperator*{\argmax}{arg\,max}

\begin{document}
\title{Model-based Reinforcement Learning: A Survey.}
\author{Thomas M. Moerland$^{1}$, Joost Broekens$^1$, Aske Plaat$^1$, and Catholijn M. Jonker$^{1,2}$ \\
$^1$ LIACS, Leiden University, The Netherlands \\
$^2$ Interactive Intelligence, TU Delft, The Netherlands 
}
\date{}
\maketitle

\tableofcontents

\abstract{Sequential decision making, commonly formalized as Markov Decision Process (MDP) optimization, is a important challenge in artificial intelligence. Two key approaches to this problem are reinforcement learning (RL) and planning. This paper presents a survey of the integration of both fields, better known as model-based reinforcement learning. Model-based RL has two main steps. First, we systematically cover approaches to dynamics model learning, including challenges like dealing with stochasticity, uncertainty, partial observability, and temporal abstraction. Second, we present a systematic categorization of planning-learning integration, including aspects like: where to start planning, what budgets to allocate to planning and real data collection, how to plan, and how to integrate planning in the learning and acting loop. After these two sections, we also discuss implicit model-based RL as an end-to-end alternative for model learning and planning, and we cover the potential benefits of model-based RL. Along the way, the survey also draws connections to several related RL fields, like hierarchical RL and transfer learning. Altogether, the survey presents a broad conceptual overview of the combination of planning and learning for MDP optimization.}

\vspace{0.5cm}

\noindent {\bf Keywords}: Model-based reinforcement learning, reinforcement learning, planning, search, Markov Decision Process, review, survey.

\section{Introduction} \label{sec_introduction}
Sequential decision making, commonly formalized as Markov Decision Process (MDP) \citep{bellman1954theory,puterman2014markov} optimization, is a key challenge in artificial intelligence. Two successful approaches to solve this problem are {\it planning} \citep{russell2016artificial,bertsekas1995dynamic} and {\it reinforcement learning} \citep{sutton2018reinforcement}. Planning and learning may actually be combined, in a field which is known as {\it model-based reinforcement learning}. We define model-based RL as: `any MDP approach that i) uses a model (known or learned) and ii) uses learning to approximate a global value or policy function'. 

\begin{figure}[!t]
  \centering
      \includegraphics[width = 0.60\textwidth]{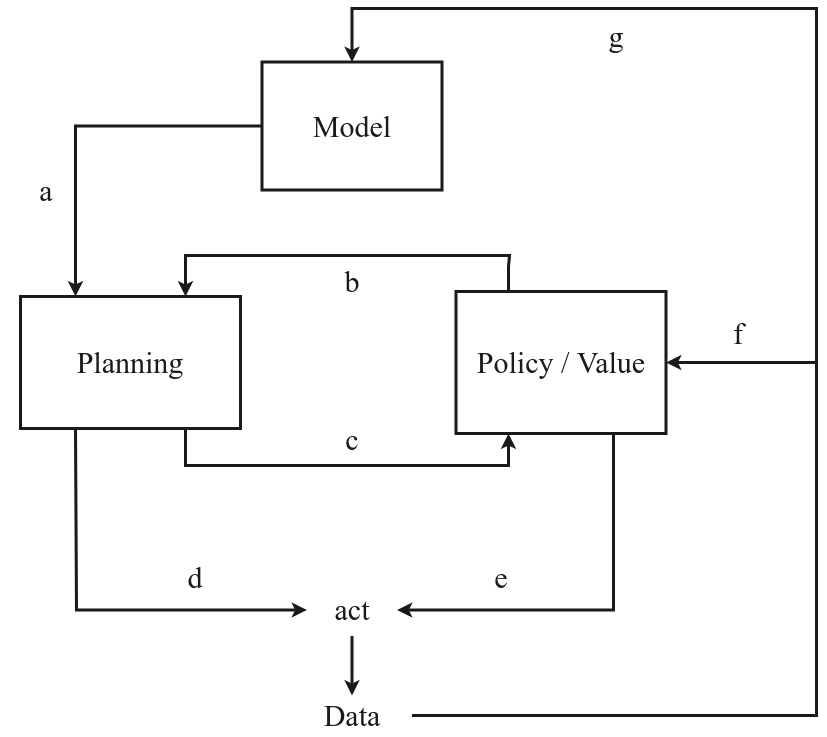}
  \caption{Overview of possible algorithmic connections between planning and learning. Learning can take place at two locations: in learning a dynamics model (arrow g), and/or in learning a policy/value function (arrows c and f). Most algorithms only implement a subset of the possible connections. Explanation of each arrow: a) plan over a learned model, b) use information from a policy/value network to improve the planning procedure, c) use the result from planning as training targets for a policy/value, d) act in the real world based on the planning outcome, e) act in the real world based on a policy/value function, f) generate training targets for the policy/value based on real world data, g) generate training targets for the model based on real world data.}
    \label{fig_model_based_integration}
\end{figure}

While model-based RL has shown great success \citep{silver2017mastering,levine2013guided,deisenroth2011pilco}, literature lacks a systematic review of the field (although \citet{hamrick2020combining} does provide an overview of mental simulation in deep learning, see Sec. \ref{sec_relatedwork} for a detailed discussion of related work). Therefore, this article presents a survey of the combination of planning and learning. A general scheme of the possible connections between planning and learning, which we will use throughout the survey, is shown in Figure \ref{fig_model_based_integration}. 

The survey is organized as follows. After a short introduction of the MDP optimization problem (Sec. \ref{sec_problem_def}), we first define the categories of model-based reinforcement learning and their relation to the fields of planning and model-free reinforcement learning (Sec. \ref{sec_categories}). Afterwards, Sections \ref{sec_model_learning}-\ref{sec_benefits} present the main body of this survey. The crucial first step of most model-based RL algorithms is {\it dynamics model learning} (Fig \ref{fig_model_based_integration}, arrow g) which we cover in Sec. \ref{sec_model_learning}. When we have obtained a model, the second step of model-based RL is to integrate planning and learning (Fig \ref{fig_model_based_integration}, arrows a-f), which we discuss in Sec. \ref{sec_model_using}. Interestingly, some model-based RL approaches do not explicitly define one or both of these steps (model learning and integration of planning and learning), but rather wrap them into a larger (end-to-end) optimization. We call these methods {\it implicit model-based RL}, which we cover in Sec. \ref{sec_implicit_mbrl}. Finally, we conclude the main part of this survey with a discussion of the potential benefits of these approaches, and of model-based RL in general (Sec. \ref{sec_benefits}). 
 
While the main focus of this survey is on the practical/empirical aspects of model-based RL, we also shortly highlight the main theoretical results on the convergence properties of model-based RL algorithms (Sec. \ref{sec_theory}). Additionally, note that model-based RL is a fundamental approach to sequential decision making, and many other sub-disciplines in RL have a close connection to model-based RL. For example, {\it hierarchical reinforcement learning} \citep{barto2003recent} can be approached in a model-based way, where the higher-level action space defines a model with temporal abstraction. Model-based RL is also an important approach to {\it transfer learning} \citep{taylor2009transfer} (through model transfer between tasks) and {\it targeted exploration} \citep{thrun1992efficient}. When applicable, the survey also presents short overviews of such related RL research directions. Finally, the survey finishes with Related Work (Sec. \ref{sec_relatedwork}), Discussion (Sec. \ref{sec_discussion}), and Summary (Sec. \ref{sec_summary}) sections. 

\section{Background} \label{sec_problem_def}
The formal definition of a {\it Markov Decision Process} (MDP) \citep{puterman2014markov} is the tuple $\{\mathcal{S},\mathcal{A},\mathcal{T},\mathcal{R},p(s_0),\gamma\}$. The environment consists of a {\it transition function} $\mathcal{T}:\mathcal{S} \times \mathcal{A} \to p(\mathcal{S})$ and a {\it reward function} $\mathcal{R}: \mathcal{S} \times \mathcal{A} \times \mathcal{S} \to \mathbb{R}$. At each timestep $t$ we observe some state $s_t \in \mathcal{S}$ and pick an action $a_t \in \mathcal{A}$. Then, the environment returns a next state $s_{t+1} \sim \mathcal{T}(\cdot|s_t,a_t)$ and associated scalar reward $r_t = \mathcal{R}(s_t,a_t,s_{t+1})$. The first state is sampled from the initial state distribution $p(s_0)$. Finally, $\gamma \in [0,1]$ denotes a discount parameter.

The agent acts in the environment according to a {\it policy} $\pi: \mathcal{S} \to p(\mathcal{A})$. In the search community, a policy is also known as a {\it contingency plan} or {\it strategy} \citep{russell2016artificial}. By repeatedly selecting actions and transitioning to a next state, we can sample a {\it trace} through the environment. The {\it cumulative return} of a trace through the environment is denoted by: $J_t = \sum_{k=0}^K \gamma^k \cdot r_{t+k}$, for a trace of length $K$. For $K=\infty$ we call this the infinite-horizon return. 

Define the action-value function $Q^\pi(s,a)$ as the expectation of the cumulative return given a certain policy $\pi$:

\begin{equation} 
Q^\pi(s,a) \dot{=} \mathbb{E}_{\pi,\mathcal{T}} \Bigg[ \sum_{k=0}^K \gamma^k r_{t+k}  \Big| s_t=s, a_t=a \Bigg]
\end{equation}

This equation can be written in a recursive form, better known as the {\it Bellman equation}:
 
\begin{align}
Q^\pi(s,a) &= \mathbb{E}_{s' \sim \mathcal{T}(\cdot|s,a)} \Bigg[ \mathcal{R}(s,a,s') + \gamma \hspace{0.1cm}\mathbb{E}_{a' \sim \pi(\cdot|s')}\Big[Q^\pi(s',a') \Big]  \Bigg] \label{eq_bellman}
\end{align}

Our goal is to find a policy $\pi$ that maximizes our expected return $Q^\pi(s,a)$:

\begin{equation} 
\pi^\star = \argmax_\pi Q^\pi(s,a) = \argmax_\pi \mathbb{E}_{\pi,\mathcal{T}} \Bigg[ \sum_{k=0}^K \gamma^k r_{t+k}  \Big| s_t=s, a_t=a \Bigg]
\end{equation}

There is {\it at least} one optimal policy, denoted by $\pi^\star$, which is better or equal than all other policies \citep{sutton2018reinforcement}. In the planning and search literature, the above problem is typically formulated as a cost {\it minimization} problem \citep{russell2016artificial}, instead of a reward maximization problem. That formulation is interchangeable with our presentation by negating the reward function.

\section{Categories of Model-based Reinforcement Learning} \label{sec_categories}
To properly define model-based reinforcement learning, we first need individual definitions of planning and reinforcement learning. There are in principle two ways to distinguish both fields: based on their access to the MDP dynamics, and based on the way they represent the solution. Regarding the first distinction, planning methods tend to have {\it reversible} access to the MDP dynamics, which allows the agent to repeatedly plan forward from the same state (similar to the way humans plan in their mind). Thereby, reversible access to the MDP dynamics allows the agent to query the MDP at any preferred state-action pair (in any preferred order). We call such reversible access to the MDP dynamics a {\it model}. 

\begin{quote}
\centering \it
A model is a form of reversible access to the MDP dynamics (known or learned).
\end{quote}

\noindent In contrast, model-free reinforcement learning approaches typically have {\it irreversible} access to the MDP dynamics, which means that the agent has to move forward from the resulting next state after executing a particular action (similar to the way we act in the real world). This essentially restricts the order in which we can visit state-action pairs in the MDP. 

Based on this different type of access to the MDP dynamics, both fields have also focused on a different type of representation of the solution. Planning methods typically use a {\it local} representation of the solution, which only stores the solution for a subset of all states (for example around a current state, which gets discarded afterwards). In contrast, learning methods cannot repeatedly plan from the same state, and therefore have to store a {\it global} solution, which stores a value function or policy for the entire state space.  

\begin{table}[!t]
\centering
\small
\caption{Distinction between planning, model-free RL, and model-based RL, based on 1) the presence of reversible access to the MDP dynamics (a model, known or learned) and 2) the presence of a global (learned) solution (e.g., policy or value function).}
\label{tab_model_based_boundaries}
\begin{tabular}{ p{6.5cm}  P{2.5cm}  P{2.5cm} }
\toprule
   & \centering \bfseries{Model} & \bfseries{Global solution} \\
\bottomrule

\it Planning &  + &  -  \\
\it Reinforcement learning & +/- &  +   \\
 
\it \quad Model-free reinforcement learning &  - &  +  \\
\it \quad Model-based reinforcement learning & + &  +    \\

\bottomrule
\end{tabular}

\end{table}

Unfortunately, the two distinctions between planning and learning (reversible versus irreversible MDP access and local versus global solution) do not always agree. For example, AlphaZero \citep{silver2018general} combines reversible access to the MDP dynamics (which would make it planning) with a global policy and value function (which would make it learning). Since many researchers consider AlphaZero a (model-based) reinforcement learning algorithm, we decide to define RL based on the presence of a global (learned) solution. This leads to the following definitions of (MDP) planning and learning (Table \ref{tab_model_based_boundaries}):

\begin{quote}
\centering \it
Planning is a class of MDP algorithms that 1) use a model and 2) store a local solution.   
\end{quote}

\begin{quote}
\centering \it
Reinforcement learning is a class of MDP algorithms that store a global solution.
\end{quote}

Starting from their different assumptions, both research fields have developed their own methodology. Along the way they started to meet, in a field that became known as {\it model-based reinforcement learning} \citep{sutton1990integrated}. The principles beneath model-based RL were also discovered in the planning community, in the form of Learning Real-Time A$^\star$ \citep{korf1990real}, while the underlying principles already date back to the Checkers programme by \citet{samuel1967some}. The key idea of model-based reinforcement learning is to combine a model and a global solution in one algorithm (Table \ref{tab_model_based_boundaries}):

\begin{quote}
\centering \it
Model-based reinforcement learning is a class of MDP algorithms that 1) use a model, and 2) store a global solution.
\end{quote}

Regarding the combination of planning and learning, `learning' is itself actually an overloaded term, since it can happen at two locations in the algorithm: 1) to learn a dynamics model, and 2) to learn a global solution (e.g., a policy or value function). Since learning can happen at two locations in algorithms that combine planning and learning, we actually end up with three subcategories of planning-learning integration (Table \ref{tab_model_based_boundaries2}):

\begin{itemize}
\item {\it Model-based RL with a learned model}, where we both learn a model and learn a global solution. An example is Dyna \citep{sutton1991dyna}.
\item {\it Model-based RL with a known model}, where we plan over a known model, and only use learning for the global solution. An example is AlphaZero \citep{silver2018general}, while also Dynamic Programming \citep{bellman1966dynamic} technically belongs to this group. 
\item {\it Planning over a learned model}, where we do learn a model, but subsequently locally plan over it, without learning a global solution. An example is Embed2Control \citep{watter2015embed}. 
\end{itemize}

Note that `planning over a learned model' is not considered model-based RL, since it does not learn a global solution to the problem (see Table \ref{tab_model_based_boundaries}). However, it is a form of planning-learning integration, and we therefore still include this topic in the survey. It is important to distinguish between the subcategories of planning-learning integration, because they also need to cope with different challenges. For example, approaches with a learned dynamics model typically need to account for model uncertainty, while approaches with a known/given dynamics model can ignore this issue, and put stronger emphasis on asymptotic performance. In the next section (Sec. \ref{sec_model_learning}) we will first discuss the various approaches to model learning (first column of Table \ref{tab_model_based_boundaries2}), while the subsequent Sec. \ref{sec_model_using} will cover the ways to integrate planning with learning of a global value or policy function (second column of Table \ref{tab_model_based_boundaries2}).

\begin{table}[!t]
\centering
\small
\caption{Categories of planning-learning integration (as covered in this survey). In MDP algorithms that use planning, learning may happen at two locations: i) to learn a model, and 2) to learn the global solution (e.g., a policy or value function). These leads to three possible combinations of planning and learning, as shown in the table.}
\label{tab_model_based_boundaries2}
\begin{tabular}{ p{6cm}  P{2.9cm}  P{2.9cm}}
\toprule
 & \centering \bfseries{Learned model} & \bfseries{Learned solution} \\
\bottomrule
\it Model-based RL with a known model &  - &   +   \\
\it Model-based RL with a learned model &  + &  +  \\
\it Planning over a learned model & + &   -   \\
\bottomrule
\end{tabular}

\end{table}


\section{Dynamics Model Learning} \label{sec_model_learning}
The first step of model-based RL (with a learned model) involves learning the dynamics model from observed data. In the control literature, dynamics model learning is better known as {\it system identification} \citep{aastrom1971system,ljung2001system}. We will first cover the general considerations of learning a one-step model (Sec. \ref{sec_approximation_methods}). Afterwards, we extensively cover the various challenges of model learning, and their possible solutions. These challenges are stochasticity (Sec. \ref{sec_stochasticity}), uncertainty due to limited data (Sec. \ref{sec_uncertainty}), partial observability (Sec. \ref{sec_partial_observability}), non-stationarity (Sec. \ref{sec_non_stationarity}),  multi-step prediction (\ref{sec_multi_step_prediction}), state abstraction (Sec. \ref{sec_state_abstraction}) and temporal abstraction (Sec. \ref{sec_action_abstraction}). The reader may wish to skip some of these section if the particular challenge is not relevant to your research problem or task of interest. 

\subsection{Basic considerations} \label{sec_approximation_methods}
Model learning is essentially a supervised learning problem \citep{jordan1992forward}, and many topics from the supervised learning community apply here. We will first focus on a simple one-step model, and discuss the three main considerations: what type of model do we learn, what type of estimation method do we use, and in what region should our model be valid? 

\begin{figure}[!t]
  \centering
      \includegraphics[width = 0.95\textwidth]{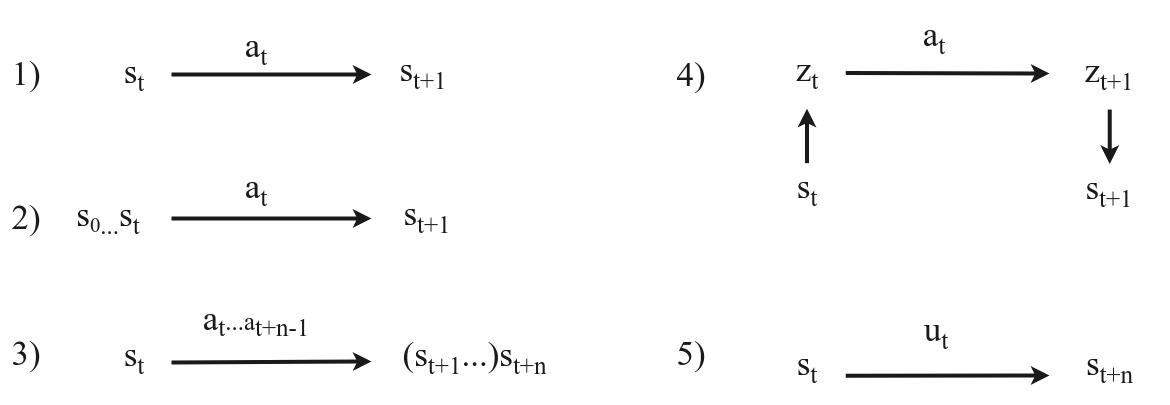}
  \caption{Overview of different types of mappings in model learning. {\bf 1)} Standard Markovian transition model $s_t,a_t \to s_{t+1}$. {\bf 2)} Partial observability (Section \ref{sec_partial_observability}). We model $s_0...s_t,a_t \to s_{t+1}$, leveraging the state history to make an accurate prediction. {\bf 3)} Multi-step prediction (Section \ref{sec_multi_step_prediction}), where we model $s_t,a_t...a_{t+n-1} \to s_{t+n}$, to predict the $n$ step effect of a sequence of actions. {\bf 4)} State abstraction (Section \ref{sec_state_abstraction}), where we compress the state into a compact representation $z_t$ and model the transition in this latent space. {\bf 5)} Temporal/action abstraction (Section \ref{sec_action_abstraction}), better known as hierarchical reinforcement learning, where we learn an abstract action $u_t$ that brings us to $s_{t+n}$. Temporal abstraction directly implies multi-step prediction, as otherwise the abstract action $u_t$ is equal to the low level action $a_t$. All the above ideas (2-5) are orthogonal and can be combined.}
    \label{fig_multi_step_prediction}
\end{figure}

\paragraph{Type of model}
We will here focus on {\it dynamics models}, which attempt to learn the transition probabilities between states. Technically, the reward function is also part of the model, but it is usually easier to learn (since we only need to predict an additional scalar in our learned dynamics function). We therefore focus on the dynamics model here. Given a batch of one-step transition data $\{s_t,a_t,r_t,s_{t+1}\}$, there are three main types of dynamics functions we might be interested in:

\begin{itemize}
\item {\it Forward model}: $(s_t,a_t) \to s_{t+1}$. This predicts the next state given a current state and chosen action (Figure \ref{fig_multi_step_prediction}, arrow 1). It is by far the most common type of model, and can be used for lookahead planning. Since model-based RL has mostly focused on forward models, the remainder of this section will primarily focus on this type of model. 
\item {\it Backward/reverse model}: $s_{t+1} \to (s_t,a_t)$. This model predicts which states are the possible precursors of a particular state. Thereby, we can plan in the backwards direction, which is for example used in prioritized sweeping \citep{moore1993prioritized}.  
\item {\it Inverse model}: $(s_t,s_{t+1}) \to a_t$. An inverse model predicts which action is needed to get from one state to another. It is for example used in RRT planning \citep{lavalle1998rapidly}. As we will later see, this function can also be useful as part of representation learning (Sec. \ref{sec_state_abstraction}). 
\end{itemize}

\paragraph{Estimation method}
We also need to determine what type of approximation method (supervised learning method) we will use. We discriminate between parametric and non-parametric methods, and between exact and approximate methods.

\begin{itemize}
\item {\it Parametric}: Parametric methods are the most popular approach for model approximation. Compared to non-parametric methods, a benefit of parametric methods is that their number of parameters is independent of the size of the observed dataset. There are two main subgroups:
 
\begin{itemize}
\item {\it Exact}: An important distinction in learning is between exact/tabular and approximate methods. For a discrete MDP (or a discretized version of a continuous MDP), a tabular method maintains a separate entry for every possible transition.  For example, in a stochastic MDP (in which we need to learn a probability distribution, see next section) a {\it tabular maximum likelihood model} \citep{sutton1991dyna} estimates the probability of each possible transition as

\begin{equation}
T(s'|s,a) = \frac{n(s,a,s')}{\sum_{s'} n(s,a,s')},
\end{equation}

where $T$ denotes the approximation of the true dynamics $\mathcal{T}$, and $n(s,a,s')$ denotes the number of times we observed $s'$ after taking action $a$ in state $s$. This approach therefore effectively normalizes the observed transition counts.	Tabular models were popular in initial model-based RL \citep{sutton1990integrated}. However, they do not scale to high-dimensional problems, as the size of the required table scales exponentially in the dimensionality of $\mathcal{S}$ (the curse of dimensionality).

\item {\it Approximate}: We may use function approximation methods, which lower the required number of parameters and allow for generalization. Function approximation is therefore the preferred approach in higher-dimensional problems. We may in principle use any parametric approximation method to learn the model. Examples include linear regression \citep{sutton2008dyna,parr2008analysis}, Dynamic Bayesian networks (DBN) \citep{hester2012learning}, nearest neighbours \citep{jong2007model}, random forests \citep{hester2013texplore}, support vector regression \citep{muller1997predicting} and neural networks \citep{werbos1989neural,narendra1990identification,wahlstrom2015pixels,oh2015action}. Especially (deep) neural networks have become popular in the last decade, for function approximation in general \citep{goodfellow2016deep}, and therefore also for dynamics approximation. Compared to the other methods, neural networks especially scale (computationally) well to high-dimensional inputs, while being able to flexibly approximate non-linear functions. Nevertheless, other approximation methods still have their use as well.
\end{itemize}

\item {\it Non-parametric}: The other main supervised learning approach is non-parametric approximation. The main property of non-parametric methods is that they directly store and use the data to represent the model. 

\begin{itemize}
\item {\it Exact}: Replay buffers \citep{lin1992self} can actually be regarded as non-parametric versions of tabular transition models, and the line between model-based RL and replay buffer methods is indeed thin \citep{vanseijen2015deeper,van2019use}.
\item {\it Approximate}: We may also apply non-parametric methods when we want to be able to generalize information to similar states. For example, Gaussian processes \citep{wang2006gaussian,deisenroth2011pilco} have been a popular non-parametric approach. Gaussian processes can also provide good uncertainty estimates, which we will further discuss in Sec. \ref{sec_uncertainty}.  

\end{itemize}
The computational complexity of non-parametric methods depends on the size of the dataset, which makes them in general less applicable to high-dimensional problems, where we usually require more data.

\end{itemize}

\noindent Throughout this work, we sometimes refer to the term `function approximation'. We then imply all {\it non-tabular} (non-exact) methods (both parametric and non-parametric), i.e., all methods that can generalize information between states.

\paragraph{Region in which the model is valid}
The third important consideration is the region of state space in which we aim to make the model valid:

\begin{itemize}
\item {\it Global}: These models approximate the dynamics over the entire state space. This is the main approach of most model learning methods.
\item {\it Local}: The other approach is to only locally approximate the dynamics, and each time discard the local model after planning over it. This approach is especially popular in the control community, where researchers frequently fit local linear approximations of the dynamics around some current state \citep{atkeson1997locally,bagnell2001autonomous,levine2014learning}. A local model restricts the input domain in which the model is valid, and is also fitted to a restricted set of data. A benefit of local models is that we may use a more restricted function approximation class (like linear), and potentially have less instability compared to global approximation. On the downside, we continuously have to estimate new models, and therefore cannot continue to learn from all collected data (since it is infeasible to store all previous datapoints).
\end{itemize}

\noindent The distinction between global and local is equally relevant for representation of a value or policy function, as we will see in Sections \ref{sec_model_using} and \ref{sec_benefits}. 

This concludes our discussion of the three basic considerations (type of model, type of estimation method, region of validity) of model learning. In practice, most model learning methods actually focus on one particular combination: a forward model, with parametric function approximation, and global coverage. The remainder of this section will discuss several more more advanced challenges of model learning, in which this particular setting will also get the most attention. 
 
\begin{figure}[!t]
  \centering
      \includegraphics[width = 0.95\textwidth]{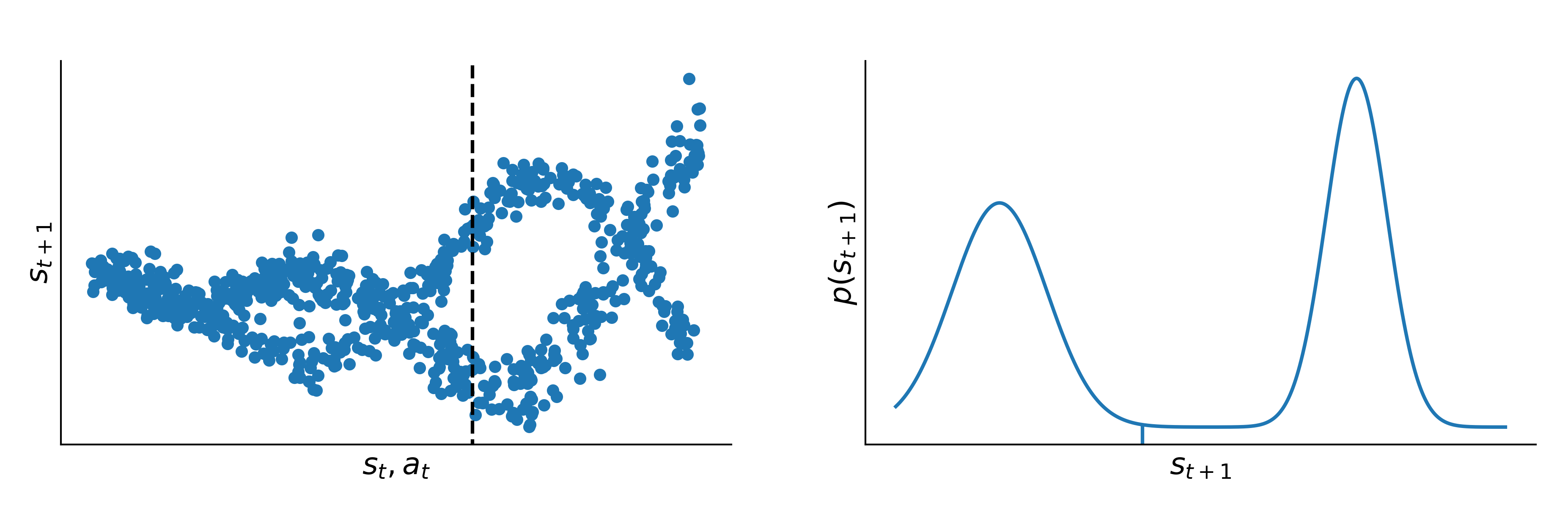}
  \caption{Illustration of stochastic transition dynamics. {\bf Left}: 500 samples from an example transition function $\mathcal{T}(s'|s,a)$. The vertical dashed line indicates the cross-section distribution on the right. {\bf Right}: distribution of $s_{t+1}$ for a particular $s,a$. We observe a multimodal distribution. The conditional mean of this distribution, which would be predicted by mean squared error (MSE) training, is shown as a vertical line.}
    \label{fig_stochasticity}
\end{figure} 
 
\subsection{Stochasticity} \label{sec_stochasticity}
In a stochastic MDP the transition function specifies a distribution over the possible next states, instead of returning a single next state (Figure \ref{fig_stochasticity}, left). In those cases, we should also specify a model that can approximate entire distributions. Otherwise, when we for example train a deterministic neural network $f_\phi(s,a)$  on a mean-squared error loss (e.g.,  \citet{oh2015action}), then the network will actually learn to predict the conditional mean of the next state distribution \citep{moerland2017learning}. This problem is illustrated in Figure \ref{fig_stochasticity}, right. 

We can either approximate the entire next state distribution (descriptive models), or approximate a model from which we can only draw samples (generative model). Descriptive models are mostly feasible in small state spaces. Examples include tabular models, Gaussian models \citep{deisenroth2011pilco} and Gaussian mixture models \citep{khansari2011learning}, where the mixture contribution typically involved {\it expectation-maximization} (EM) style inference \citep{ghahramani1999learning}. However, these methods do not scale well to high-dimensional state spaces. 

In high-dimensional problems, most successful attempts are based on neural network approximation (deep generative models). One approach is to use variational inference (VI) to estimate dynamics models \citep{depeweg2016learning,moerland2017learning,babaeizadeh2017stochastic,buesing2018learning}. Competing approaches include generative adversarial networks (GANs), autoregressive full-likelihood models, and flow-based density models, which were applied to sequence modeling by \citet{yu2017seqgan}, \citet{kalchbrenner2017video} and \citet{ziegler2019latent}, respectively. Detailed discussion of these methods falls outside the scope of this survey, but there is no clear consensus yet which deep generative modeling approach works best.

\subsection{Uncertainty} \label{sec_uncertainty}
\begin{figure}[!t]
  \centering
      \includegraphics[width = 0.95\textwidth]{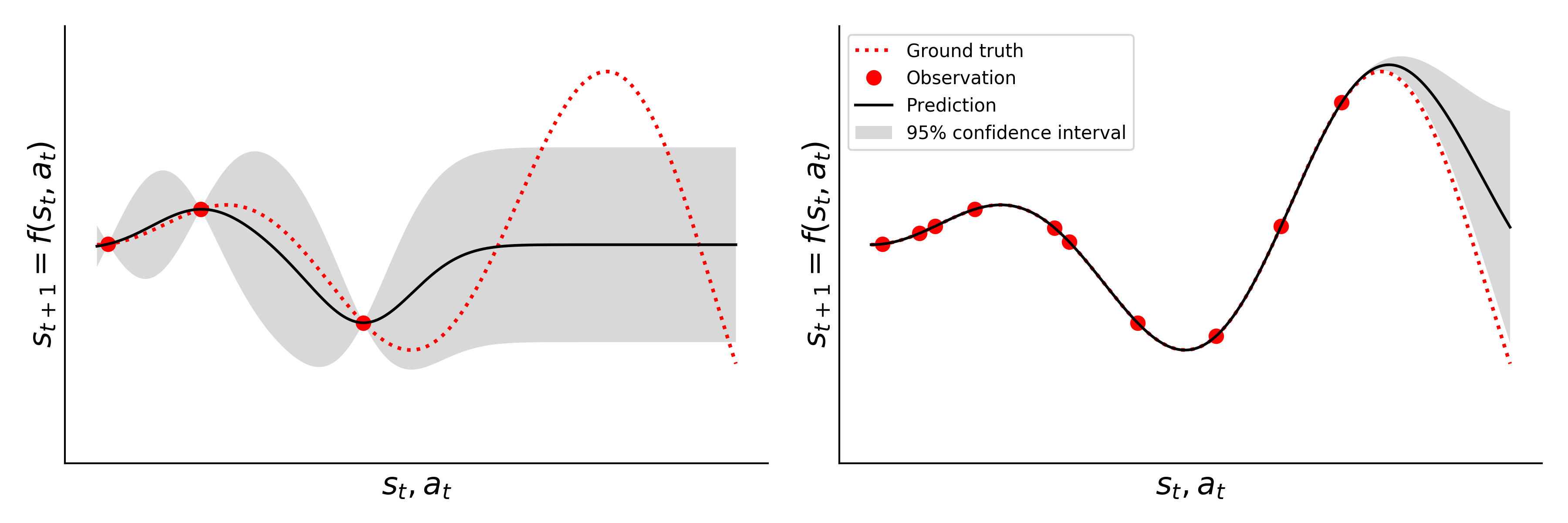}
  \caption{Illustration of uncertainty due to limited data. Red dotted line depicts an example ground truth transition function. {\bf Left}: Gaussian Process fit after 3 observations. The predictions are clearly off in the right part of the figure, due to wrong extrapolation. The shaded area shows the 95\% confidence interval, which does identify the remaining uncertainty, although not completely correct. {\bf Right}: Gaussian Process fit after 10 observations. Predictions are much more certain now, mostly matching the true function. There is some remaining uncertainty on the far right of the curve. }
    \label{fig_uncertainty}
\end{figure}

A crucial challenge of model-based learning is dealing with uncertainty due to limited data. Uncertainty due to limited data (also known as {\it epistemic} uncertainty) clearly differs from the previously discussed stochasticity (also known as {\it aleatoric} uncertainty) \citep{der2009aleatory}, in the sense that epistemic uncertainty can be reduced by observing more data, while stochasticity can never be reduced. We will here focus on methods to estimate (epistemic) uncertainty, which is an important topic in model-based RL, since we need to assess whether our plan is actually reliable. Note that uncertainty is even relevant in the absence of stochasticity, as illustrated in Figure \ref{fig_uncertainty}. 

We therefore want to estimate the uncertainty around our predictions. Then, when we plan over our model, we can detect when our predictions become less trustworthy. There are two principled approaches to uncertainty estimation in statistics: frequentist and Bayesian. A frequentist approach is for example the statistical bootstrap, applied to model estimation by \citet{frohlich2014uncertainty} and \citet{chua2018deep}. Bayesian RL methods were previously surveyed by \citet{ghavamzadeh2015bayesian}. Especially successful have been non-parametric Bayesian methods like Gaussian Processes (GPs), for example used for model estimation in PILCO \citep{deisenroth2011pilco}. However, GPs scale (computationally) poorly to high-dimensional state spaces. Therefore, there has been much recent interest in Bayesian methods for neural network approximation of dynamics, for example based on variational dropout \citep{gal2016improving} and variational inference \citep{depeweg2016learning}. Note that uncertainty estimation is also an active research topic in the deep learning community itself, and advances in those fields will likely benefit model-based RL as well. We will discuss how to plan over an uncertain model in Sec. \ref{sec_model_using}. 

\subsection{Partial observability} \label{sec_partial_observability}
Partial observability occurs in an MDP when the current observation does not provide all information about the ground truth state of the MDP. Note the difference between partial observability and stochasticity. Stochasticity is fundamental noise in the transition of the ground truth state, and can not be mitigated. Instead, partial observability originates from a lack of information in the current observation, but can partially be mitigated by incorporating information from previous observations (Figure \ref{fig_multi_step_prediction}, arrow 2). For example, a first-person view agent can not see what is behind itself right now, but it can remember what it saw behind itself a few observations ago, which mitigates the partial observability.

So how do we incorporate information from previous observations? We largely identify four approaches: i) windowing, ii) belief states, iii) recurrency and iv) external memory (Figure \ref{fig_partial_observability}). 

\begin{itemize}
\item {\it Windowing}: In the windowing approach we concatenate the $n$ most recent observations and treat these together as the state \citep{lin1992memory}. \citet{mccallum1997reinforcement} extensively studies how to adaptively adjust the window size. In some sense, this is a tabular solution to partial observability. Although this approach can be effective in small problems, they suffer from high memory requirements in bigger problems, and cannot profit from generalization.

\item {\it Belief states}: Belief states explicitly partition the learned dynamics model in an {\it observation model} $p(o|s)$ and a {\it latent transition model} $\mathcal{T}(s'|s,a)$ \citep{chrisman1992reinforcement}. This structure is similar to the sequence modeling approach of {\it state-space models} \citep{bishop2006pattern}, such as hidden Markov models (HMM). This approach represents the dynamics model as a probabilistic graph, in which the parameters are for example estimated through expectation-maximization (EM) \citep{ghahramani1996parameter}. There is also specific literature on planning for such belief state models, known as POMDP planners \citep{spaan2004point,kurniawati2008sarsop,silver2010monte}. The principles of belief state models have also been combined with neural networks \citep{krishnan2015DeepKF,karl2016deep}, which makes them applicable to high-dimensional problems as well. 

\begin{figure}[!t]
  \centering
      \includegraphics[width = 0.95\textwidth]{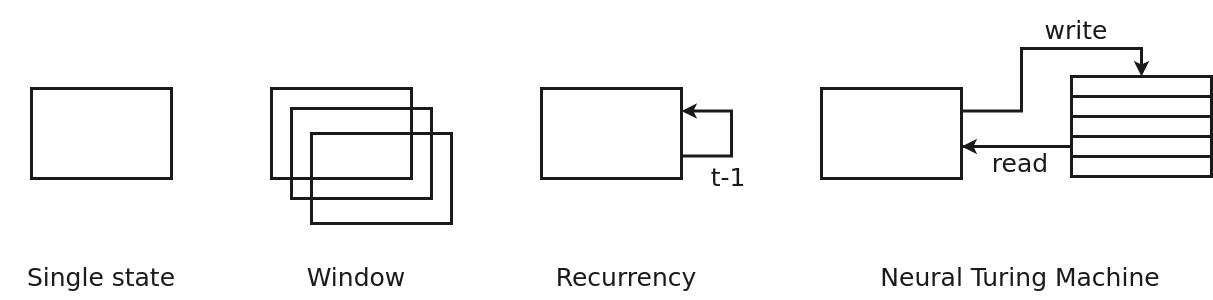}
  \caption{Example approaches to partial observability. The window approach concatenates the most recent $n$ frames and treats this as a new state. The recurrent approach learns a recurrent mapping between timesteps to propagate information. The Neural Turing Machine uses an external memory to explicitly write away information and read it back when relevant, which is especially applicable to long-range dependencies.}
    \label{fig_partial_observability}
\end{figure} 

\item {\it Recurrency}: The most popular solution to partial observability is probably the use of {\it recurrent} neural networks, first applied to dynamics learning in \citet{lin1993reinforcement,parlos1994application}. A variety of papers have studied RNNs in high-dimensional settings in recent years \citep{chiappa2017recurrent,ha2018recurrent,gemici2017generative}. Since the transition parameters of the RNN are shared between all timesteps, the model size is independent of the history length, which is one the main benefits of RNNs. They also neatly integrate with gradient-based training and high-dimensional state spaces. However, they do suffer from vanishing and exploding gradients to model long-range dependencies. This may be partly mitigated by long short-term memory (LSTM) cells \citep{hochreiter1997long} or temporal skip connections \citep{el1996hierarchical}. \citet{beck2020amrl} recent proposed {\it aggregators}, which are more robust to long-range stochasticity in the observed sequences, as frequently present in RL tasks.

\item {\it External memory}: The final approach to partial observability is the use of an external memory. \citet{peshkin1999learning} already gave the agent access to arbitrary bits in its state that could be flipped by the agent. Over time it learned to correctly flip these bits to memorize historical information. A more flexible extension of this idea are Neural Turing Machines (NTM) \citep{graves2014neural}, which have read/write access to an external memory, and can be trained with gradient descent. \citet{gemici2017generative} study NTMs in the context of model learning. External memory is especially useful for long-range dependencies, since we do not need to keep propagating information, but can simply recall it once it becomes relevant. The best way to store and recall information is however still an open area of research. 
\end{itemize}

Partial observability is an inherent property of nearly all real-world tasks. When we ignore partial observability, our solution may completely fail. Therefore, many research papers that focus on some other research question, still incorporate methodology to battle the partial observability in the domain. Note that the above partial observability methodology is equally applicable to a learned policy or value function.

\subsection{Non-stationarity} \label{sec_non_stationarity}
\begin{figure}[t]
  \centering
      \includegraphics[width = 0.95\textwidth]{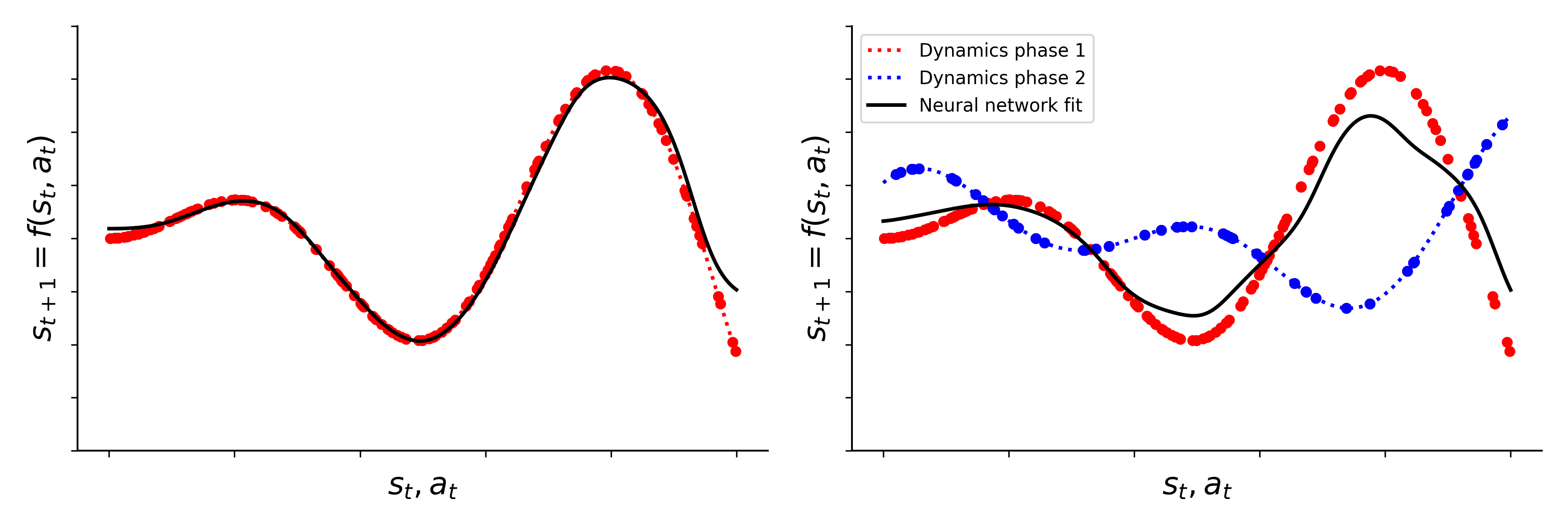}
  \caption{Illustration of non-stationarity. {\bf Left}: First 150 data points sampled from initial dynamics. Black line shows the prediction of a neural network with 2 hidden layers of 50 units and tanh activations trained for 150 epochs. {\bf Right}: Due to non-stationarity the dynamics changed to the blue curve, from which we sample an additional 50 points. The black curve shows the new neural network fit without detection of the dynamics change, i.e., treating all data as valid samples from the same transition distribution. We clearly see the network has trouble adapting to the new regime, as it still tries to fit to the old dynamics data points as well.}
    \label{fig_nonstationarity}
\end{figure}

Non-stationarity in an MDP occurs when the true transition and/or reward function change(s) over time. When the agent keeps trusting its previous model, without detecting the change, then its performance may deteriorate fast (see Figure \ref{fig_nonstationarity} for an illustration of the problem). The main approach to non-stationarity are {\it partial models} \citep{doya2002multiple}. Partial models are an ensemble of stationary models, where the agent tries to detect a regime switch to subsequently switch between models as well. \citet{da2006dealing} detect a switch based on the prediction errors in transition and reward models, while \citet{nagabandi2018deep} makes a soft assignment based on a Dirichlet process. \citet{jaulmes2005learning} propose a simpler approach than partial models, by simply strongly decaying the contribution of older data (which is similar to a high learning rate). However, a higher learning rate may also make training unstable.

\subsection{Multi-step Prediction} \label{sec_multi_step_prediction}
The models we discussed so far made one-step predictions of the next state. However, we eventually intend to use these in models in multi-step planning procedures. Of course, we can make multi-step predictions with one-step models by repeatedly feeding the prediction back into the learned model. However, since our learned model was never optimized to make long-range predictions, accumulating errors may actually cause our multi-step predictions to diverge from the true dynamics. Several authors have identified this problem \citep{talvitie2014model,venkatraman2015improving,talvitie2017self,machado2018revisiting}.

We therefore require models that are robust at long range predictions (Figure \ref{fig_multi_step_prediction}, arrow 3). There are largely two approaches to this challenge: i) different loss functions and ii) separate dynamics functions for 1,2..$n$-step predictions. In the first approach we simply include multi-step prediction losses in the overall training target \citep{abbeel2005learning,chiappa2017recurrent,hafner2018learning,ke2019learning}. These models still make 1-step predictions, but during training they are unrolled for $n$ steps and trained on a loss with the ground truth $n$-step observation. The second solution is to learn a specific dynamics model for every $n$-step prediction \citep{asadi2018towards}. In that case, we learn for example a specific function $T^3(\hat{s}_{t+3}|s_t,a_t,a_{t+1},a_{t+2})$, which makes a three step prediction conditioned on the current state and future action sequence. Some authors directly predict entire trajectories, which combines predictions of multiple depths \citep{mishra2017prediction}. The second approach will likely have more parameters to train, but prevents the instability of feeding an intermediate prediction back into the model.

Some papers do not explicitly specify how many steps in the future to predict \citep{neitz2018adaptive}, but for example automatically adjust this based on the certainty of the predicted state \citep{jayaraman2018time}. The topic of multi-step prediction also raises a question about performance measures. If our ultimate goal is multi-step planning, then one-step prediction errors are likely not a good measure of model performance.

\subsection{State abstraction} \label{sec_state_abstraction}
Representation learning is a crucial topic in reinforcement learning and control \citep{lesort2018state}. Good representations are essential for good next state predictions, and equally important for good policy and value functions. Representation learning is an important research field in machine learning itself, and many advances in state abstraction for model estimation actually build on results in the broader representation learning community. 

Early application of representation learning in RL include (soft) state aggregation \citep{singh1995reinforcement} and principal component analysis (PCA) \citep{nouri2010dimension}. \citet{mahadevan2003learning} covers various approaches to learning basis functions in Markov Decision Processes. However, by far the most successful approach to representation learning in recent years have been deep neural networks, with a variety of example applications to model learning \citep{oh2015action,watter2015embed,chiappa2017recurrent}. 

In the neural network-based approach, the dynamics model is typically factorized into three parts: i) an {\it encoding} function $z_t = f_\phi^\text{enc}(s_t)$, which maps the observation to a latent representation $z_t$, ii) a {\it latent dynamics} function $z_{t+1} = f_\phi^\text{trans}(z_t,a_t)$, which transitions to the next latent state based on the chosen action, and iii) a {\it decoder} function $s_{t+1}=f_\phi^\text{dec}(z_{t+1})$, which maps the latent state back to the next state prediction. This structure, visualized in Figure \ref{fig_multi_step_prediction}, arrow 4, reminds of an auto-encoder (with added latent dynamics), as frequently used for representation learning in the deep learning community.

There are three important additional themes for state representation learning in dynamics models: i) how do we ensure that we can plan at the more abstract level, ii) how may we better structure our models to emphasize objects and their physical interactions, and iii) how may we construct loss functions that retrieve more informative representations. 

\paragraph{Planning at a latent level}
We ideally want to be able to plan at a latent level, since it allows for faster planning. Since the representation space is usually smaller than the observation space, this may save much computational effort. However, we must ensure that the predicted next latent state lives in the same embedding space as the encoded current latent state. Otherwise, repeatedly feeding the latent prediction into the latent dynamics model will lead to predictions that diverge from the truth. One approach is to add an additional loss that enforces the next state prediction to be close to the encoding of the true next state \citep{watter2015embed}. An alternative are deep state-space models, like deep Kalman filters \citep{krishnan2015DeepKF} or deep variational Bayes filters \citep{karl2016deep}. These require probabilistic inference of the latent space, but do automatically allow for latent level planning. 

We may also put additional restrictions on the latent level dynamics that allow for specific planning routines. For example, (iterative) linear-quadratic regulator (LQR) \citep{todorov2005generalized} planning requires a linear dynamics function. Several authors \citep{watter2015embed,van2016stable,fraccaro2017disentangled} linearize their learned model on the latent level, and subsequently apply iLQR to solve for a policy \citep{watter2015embed,zhang2019solar,van2016stable}. In this way, the learned representations may actually simplify planning, although it does require that the true dynamics can be linearly represented at latent level.

State abstraction is also related to {\it grey-box} system identification. In system identification \citep{aastrom1971system}, the control term for model learning, we may discriminate `black box' and `grey box' approaches \citep{ljung2001system}. Black box methods, which do not assume any task-specific knowledge in their learning approach, are the main topic of Section \ref{sec_model_learning}. Grey box methods do partially embed task-specific knowledge in the model, and estimate remaining free parameters from data. The prior knowledge of grey box models is usually derived from the rules of physics. One may use the same idea to learn state abstractions. For example, in a robots task with visual observations, we may known the required (latent) transition model (i.e., $f_\phi^\text{trans}$ is known from physics), but not the encoding function from the visual observations ($f_\phi^\text{enc}$ is unknown). \citet{wu2015galileo} give an example of this approach, where the latent level dynamics are given by a known, differentiable physics engine, and we optimize for the encoding function from image observations.

\paragraph{Objects}
A second popular approach to improve representations is by focusing on {\it objects} and their interactions. Infants are able to track objects at early infancy, and the ability to reason about object interaction is indeed considered a core aspect of human cognition \citep{spelke2007core}. In the context of RL, these ideas have been formulated as object-oriented MDPs \citep{diuk2008object} and relational MDPs \citep{guestrin2003generalizing}. Compared to models that predict raw pixels, such object-oriented models may better generalize to new, unseen environments, since they disentangle the physics rules about objects and their interactions. 

We face two important challenges to learn an object-oriented model: 1) how do we identify objects, and 2) how do we model interaction between objects at a latent level. Regarding the first questions, several methods have provided explicit object recognizers in advance  \citep{fragkiadaki2015learning,kansky2017schema}, but other recent papers manage to learn them from the raw observations in a fully unsupervised way \citep{van2018relational,xu2019unsupervised,watters2019cobra}. The interaction between objects is typically modeled like a {\it graph neural network}. In these networks, the nodes should capture object features (e.g., appearance, location, velocity) and the edge update functions predict the effect of an interaction between two objects \citep{van2018relational}. There is a variety of recent successful examples in this direction, like Schema Networks \citep{kansky2017schema}, Interaction Networks \citep{battaglia2016interaction}, Neural Physics Engine \citep{chang2016compositional}, Structured World Models \citep{kipf2020contrastive} and COBRA \citep{watters2019cobra}. In short, object-oriented approaches tend to embed (graph) priors into the latent neural network structure that enforce the model to extract objects and their interactions. We refer the reader to \citet{battaglia2018relational} for a broader discussion of relational world models.

\paragraph{Better loss functions}
Another way to achieve more informative representations is by constructing better loss functions. First of all, we may share the representation layers of the model with other prediction tasks, like predicting the reward function. The idea to share different prediction targets to speed-up representation learning is better known as an `auxilliary loss' \citep{jaderberg2016reinforcement}.

We may also construct other losses for which we do not directly observe the raw target. For example, a popular approach is to predict the {\it relative} effect of actions: $s_{t+1} - s_t$ \citep{finn2016unsupervised}. Such background subtraction ensures that we focus on moving objects. An extension of this idea is {\it contingency awareness}, which describes the ability to discriminate between environment factors within and outside our control \citep{watson1966development}. We would also like to emphasize these controllable aspects in our representations. One way to achieve this is through an inverse dynamics loss, where we try to predict the action that achieves a certain transition: ($s,s') \to a$ \citep{pathak2017curiosity}. This will focus on those parts of the state that the chosen action affects. Other approaches that emphasize controllable factors can be found in \citet{choi2018contingency,thomas2018disentangling,sawada2018disentangling}. 

There is another important research line which improves representations through {\it contrastive} losses. A contrastive loss is not based on a single data point, but on the similarity or dissimilarity with other observations. As an example, \citet{sermanet2018time} record the same action sequence from different viewpoints, and obtains a compact representation by enforcing similar states from different viewpoints to be close to eachother in embedding space. \citet{ghosh2018learning} add a loss based on the number of actions needed to travel between states, which enforces states that are dynamically close to be close in representation space as well. This is an interesting idea, since we use representation learning to actually make planning easier. Contrastive losses have also been constructed from the rules of physics in robotics tasks \citep{jonschkowski2015learning}, have been applied to Atari models \citep{anand2019unsupervised}, and have combined with the above object-oriented approach \citep{kipf2020contrastive}.

Finally, there is an additional way to improve representations through {\it value equivalent models} \citep{grimm2020value}. These models are trained on their ability to predict a value or (optimal) action. We will cover this idea in Sec. \ref{sec_implicit_mbrl} on implicit model-based RL, which covers methods that optimize elements of the model-based RL process for the ability to output an (optimal) action or value. To summarize, this section discussed the several ways in which the state representation learning of models may be improved, for example by embedding specific substructure in the networks (e.g., to extract objects and their interactions), or by constructing smarter loss functions. All these topics have their individual benefit, and future work may actually focus on ways to combine these approaches. 

\begin{figure}[!t]
  \centering
      \includegraphics[width = 0.75\textwidth]{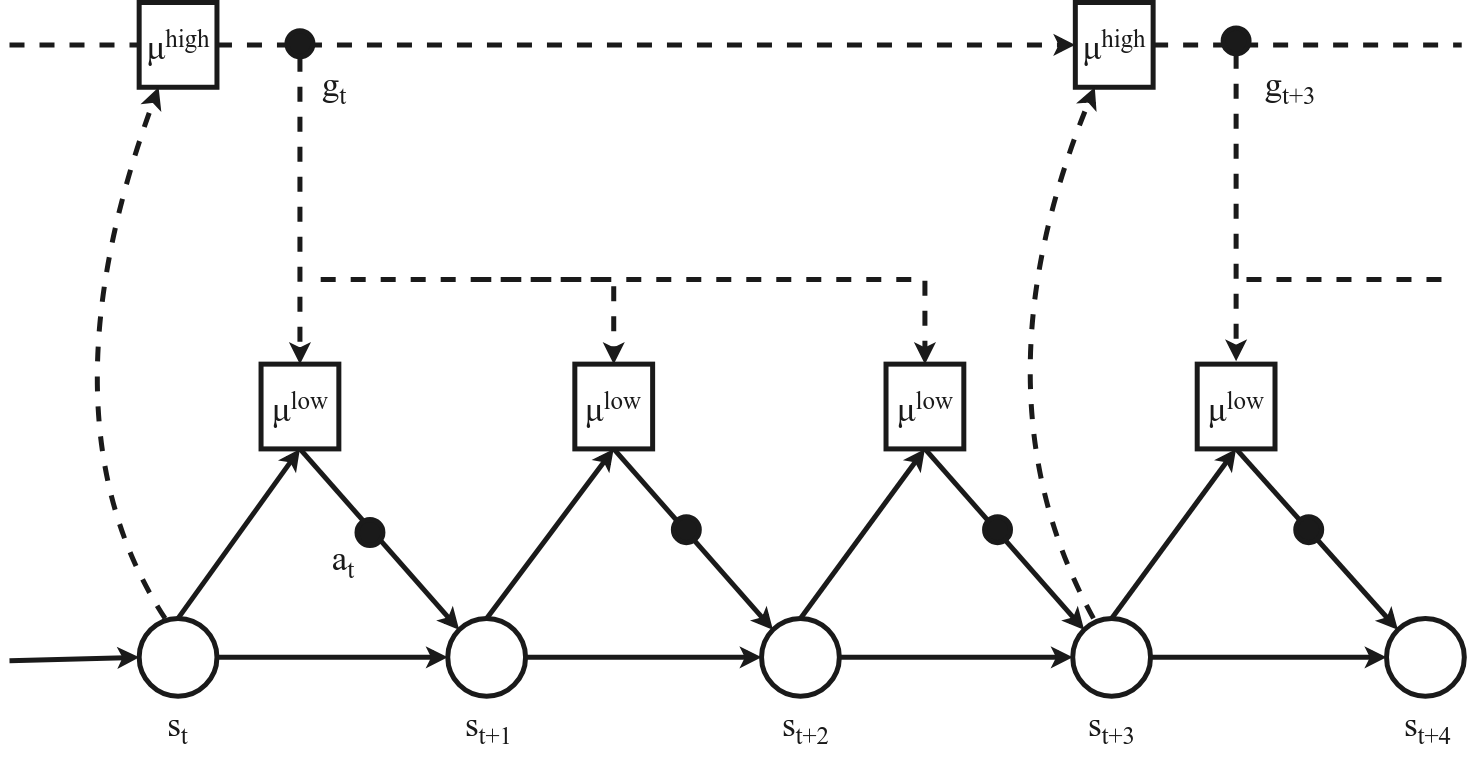}
  \caption{Conceptual illustration of a two-level hierarchy, partially based on \citet{nachum2018data}. Standard low-level interaction is shown with solid lines, temporal abstraction is shown with dashed lines. The high-level controller picks a high-level action (goal) $g_t$ according to $\pi^\text{high}$. After fixing $g_t$, the low level controller executes the relevant subpolicy, for example in the form of a goal-conditioned policy $\pi^\text{low}(s,g)$. The number of steps between high-level actions can be fixed or variable, depending on the framework. The illustration assumes full observability, in which case we only need to condition $\pi^\text{high}$ on the current observation. We may also feed $g$ back into the next high-level decision to enable temporal correlation between goals.}
    \label{fig_hierarchy}
\end{figure}  

\subsection{Temporal abstraction} \label{sec_action_abstraction}
The MDP definition typically involves low-level, atomic actions executed at a high-frequency. This generates deep search trees with long-range credit assignment. However, many of these paths give the same end-state, and some end-states are more useful than others. The idea of temporal abstraction, better known as {\it hierarchical} reinforcement learning \citep{barto2003recent,hengst2017hierarchical,thrun1995finding}, is to identify a high-level action space that extends over multiple timesteps (Figure \ref{fig_multi_step_prediction}, arrow 5 and Figure \ref{fig_hierarchy}). Indeed, temporal abstraction can theoretically reduce both the sample  \citep{brunskill2014pac} and computational complexity \citep{mann2014scaling} of solving the MDP.

There are a variety of frameworks to define abstract actions. One popular choice is the {\it options} framework \citep{sutton1999between}. Options are a discrete set of high-level actions. Each option $u$ has its own initiation set $I^u \in \mathcal{S}$ from which the option can be started, a sub-policy $\pi^u$ for execution, and a state-dependent termination probability $\beta^u(s)$ for the option to end in a reached state. A popular competing approach are {\it goal-conditioned policy/value functions} (GCVF), also known as universal value function approximators \citep{schaul2015universal}. These ideas originally date back to work on Feudal RL \citep{dayan1993feudal}. GCVFs use a goal space $\mathcal{G}$ as the abstract action space. They learn a goal-conditioned value function $Q(s,a,g)$, which estimates the value of $a$ in $s$ if we attempt to reach $g$. We train such models on a {\it goal-parametrized reward function}, which for example rewards the agent for getting closer to $g$ in Euclidean distance \citep{nachum2018data}. Afterwards, we can plan by chaining multiple subgoals \citep{eysenbach2019search,zhang2021world}. 

Options and goal-conditioned value functions are conceptually different. Most importantly, options have a separate sub-policy per option, while GCVFs attempt to generalize over goals/subpolicies. Moreover, options fix the initiation and termination set based on state information, while GCVFs can initiate and terminate everywhere. Note that GCVFs can some sense interpolate from one-step models (plan a next subgoal which is only one step away) to model-free RL (directly give the final goal to the GCVF, without any planning), as for example shown by \citet{pong2018temporal}.

\paragraph{Discovery of relevant sub-routines}

Whether we use options, GCVFs, or some other definition of abstract actions, the most important question is often: how do we actually identify the {\it relevant} subroutines, i.e., relevant end-states for our options, or goal states for our GCVF. We summarize the most important approaches below: 

\begin{itemize} 

\item {\it Graph structure}: This approach identifies `bottleneck' states as end-points for the subroutines. A bottleneck is a state that connects two densely interconnected subgraphs in the MDP graph \citep{menache2002q}. Therefore, a bottleneck is a crucial state in order to reach another region of the MDP, and therefore a candidate subgoal. There are several ways to identify bottlenecks: \citet{mcgovern2001automatic} identify bottlenecks from overlapping states in successful trials, \citet{csimcsek2005identifying} run a graph partitioning algorithm on a reconstruction of the MDP graph, and \citet{goel2003subgoal} search for states with many predecessors, but whose successors do not have many predecessors. The bottleneck approach received much attention in smaller problems, but does not scale well to higher-dimensional problems.

\item {\it State-space coverage}: Another idea is to spread the end-states of subroutines over the entire state-space, in order to reach good coverage. Most approaches first cluster the state space, and subsequently learn a dynamics model to move between the cluster centers \citep{mannor2004dynamic,lakshminarayanan2016option,machado2017laplacian}. Instead of the raw state space, we may also cluster in a compressed representation of it \citep{ghosh2018learning} (see previous section as well). 

\item {\it Compression (information-theoretic)}: We may also attempt to simply compress the space of possible end-points. This idea is close to the state space coverage ideas above. \citet{gregor2016variational,eysenbach2018diversity,achiam2018variational} associate the distribution of observed end-states with a noise distribution. After training, the noise distribution acts as a high-level action space from which we can sample. Various approaches also include additional information-theoretic regularization of this compression. For example, \citet{gregor2016variational} add the criterion that action sequences in the compressed space should make the resulting state well predictable (`empowerment'). Other examples are provided by \citet{florensa2017stochastic,hausman2018learning,fox2016principled}.

\item {\it Reward relevancy}: The idea of this approach is that relevant subroutines will help incur extra reward, and they should therefore automatically emerge from a black-box optimization approach. These approaches embed the structure of subroutines into their algorithms, ensure that the overall model is differentiable, and run an end-to-end optimization. Examples are the Option-Critic \citep{bacon2017option,riemer2018learning} and Feudal Networks \citep{vezhnevets2017feudal}, with more examples in \citet{frans2018meta,levy2018hierarchical,heess2016learning,nachum2018data}. \citet{daniel2016probabilistic,fox2017multi} use probabilistic inference based on expectation-maximization, where the E-step infers which options are active, and the M-step maximizes with respect to the value. A challenge for end-to-end approaches is to prevent degeneration, i.e., preventing that a single subroutine starts to solve the entire task, or that every subroutine terminates after one step. 

\item {\it Priors:} Finally, we may also use prior knowledge to identify useful subroutines. Sometimes, the prior knowledge is domain-specific, like pre-training on hand-coded sub-tasks \citep{tessler2017deep,heess2016learning}. \citet{kulkarni2016hierarchical} identify all objects in the scene as end-points, which may generalize over domains when combined with a generic object recognizer. Several papers also infer relevant subroutines from expert demonstrations \citep{konidaris2012robot,fox2017multi,hamidi2015active}, which is of course also a form of prior knowledge.

\end{itemize}

This concludes our discussion of temporal abstraction, and thereby also our discussion of model learning as a whole. In summary, we have seen a variety of important challenges in model learning, which several research papers have addressed. In all directions important progress has been made, but most papers tend to focus on a specific problem in isolation. Since complex real-world tasks likely require many of the discussed aspects, the combination of these methods in more complex tasks seems an important future research direction.


\section{Integration of Planning and Learning} \label{sec_model_using}

The importance of models for intelligence has been long recognized in various research fields: machine learning \citep{bellman1966dynamic,jordan1992forward}, neuroscience \citep{tolman1948cognitive,doll2012ubiquity} and behavioural psychology \citep{craik1943nature,wolpert1995internal,doll2012ubiquity}. In this section we will discuss the integration of planning and learning to arrive at a policy $\pi(a|s)$, i.e., a local or global specification of action prioritization. We will specify a taxonomy that disentangles four central questions of the integration of planning and learning. The four main questions we need to answer are: 

\begin{table}[t]
\centering
\footnotesize
\caption{Overview of taxonomy of planning-learning integration. These considerations are discussed throughout Sec. \ref{sec_model_using}. Table \ref{table_overview4} summarizes several model-based RL algorithms on these dimensions. \label{table_framework4}}
\begin{adjustbox}{width=\textwidth}
\begin{tabular}{ p{2.5cm} p{3.3cm}  p{6.0cm}  }
\toprule
\bfseries{Dimension} \newline  &  \bfseries{Consideration} &  \bfseries{Choices}  \newline \\
  \hline			
\hspace{0.1cm} \newline 1. Start state (\ref{sec_startplanning})  & 
\hspace{0.1cm} \newline - Start state & 
\hspace{0.1cm} \newline Uniform $\leftrightarrow$ visited $\leftrightarrow$ prioritized $\leftrightarrow$ current  \\

\hspace{0.1cm} \newline 2. Budget (\ref{sec_tradeoff})   & 
\hspace{0.1cm} \newline - Number of real steps before planning &
\hspace{0.1cm} \newline 1 $\leftrightarrow$ $n$, episode, etc.  \\
\hspace{0.1cm} \newline  &
\hspace{0.1cm} \newline - Effort per planning cycle &  
\hspace{0.1cm} \newline 1 $\leftrightarrow$ $n$ $\leftrightarrow$ convergence \\ 

\hspace{0.1cm} \newline 3. Planning approach (\ref{sec_howtoplan})   & 
\hspace{0.1cm} \newline - Type  &
\hspace{0.1cm} \newline Discrete $\leftrightarrow$ gradient-based \\
\hspace{0.1cm} \newline  &
\hspace{0.1cm} \newline - Direction &  
\hspace{0.1cm} \newline Forward $\leftrightarrow$ Backward \\ 
\hspace{0.1cm} \newline  &
\hspace{0.1cm} \newline - Breadth &  
\hspace{0.1cm} \newline 1 $\leftrightarrow$ adaptive $\leftrightarrow$ full \\
\hspace{0.1cm} \newline  &
\hspace{0.1cm} \newline - Depth &  
\hspace{0.1cm} \newline 1 $\leftrightarrow$ interm./adaptive $\leftrightarrow$ full \\
\hspace{0.1cm} \newline  &

\hspace{0.1cm} \newline - Uncertainty &  
\hspace{0.1cm} \newline Data-close $\leftrightarrow$ Uncertainty propagation \newline (-Prop.method: parametric $\leftrightarrow$ sample)  \\

\hspace{0.1cm} \newline 4. Integration in learning loop (\ref{sec_integration})  & 
\hspace{0.1cm} \newline - Planning input from learned function &
\hspace{0.1cm} \newline Yes (value/policy) $\leftrightarrow$ No  \\

\hspace{0.1cm} \newline    & 
\hspace{0.1cm} \newline - Planning output for training targets &
\hspace{0.1cm} \newline Yes (value/Policy) $\leftrightarrow$ No  \\
\hspace{0.1cm} \newline  &
\hspace{0.1cm} \newline - Planning output for action selection &  
\hspace{0.1cm} \newline Yes $\leftrightarrow$ No  \\

& & \\
\bottomrule
\end{tabular}
\end{adjustbox}
\end{table}

\begin{enumerate}
\item At which state do we start planning? (Sec. \ref{sec_startplanning})
\item How much planning budget do we allocate for planning and real data collection? (Sec. \ref{sec_tradeoff})
\item How to plan? (Sec. \ref{sec_howtoplan})
\item How to integrate planning in the learning and acting loop? (Sec. \ref{sec_integration})
\end{enumerate}

The first three questions mostly focus on the planning method itself (Fig. \ref{fig_model_based_integration}, arrow a), while the last question covers the way planning is integrated in the learning and acting loop (Fig. \ref{fig_model_based_integration}, arrows b-g). Note that each of the above questions actually have several important subconsiderations, which leads to the overall taxonomy summarized in Table \ref{table_framework4}. The next sections will discuss each of these subconsiderations. 

\subsection{At which state do we start planning?} \label{sec_startplanning}
The natural first question of planning is: at which state do we start? There are several options:
\begin{itemize}
\item {\it Uniform}: A straightforward approach is to uniformly select states throughout the state space. This is for example the approach of Dynamic Programming \citep{bellman1966dynamic}, which selects all possible states in a sweep. The major drawback of this approach is that it does not scale to high dimensional problems, since the total number of states grows exponentially in the dimensionality of the state space. The problem is that we will likely update many states that are not even reachable from the start state.  

\item {\it Visited}: We may ensure that we only plan at reachable states by selecting previously visited states as starting points. This approach is for example chosen by Dyna \citep{sutton1990integrated}. 

\item {\it Prioritized}: Sometimes, we may be able to obtain an ordering over the reachable states, identifying their relevancy for a next planning update. A good example is Prioritized Sweeping \citep{moore1993prioritized}, which identifies states that likely need updating. Prioritization has also been used in replay database approaches \citep{schaul2016prioritized}. 

\item {\it Current}: Finally, a common approach is to only spend planning effort at the current state of the real environment. This puts emphasis at finding a better solution or more information in the region where we are currently operating. Even model-based RL methods with a known model, like AlphaGo Zero \citep{silver2017mastering}, sometimes (because of the problem size) introduce the notion of a real environment and current state. The real environment step introduces a form of pruning, as it ensures that we move forward at some point, obtaining information about deeper nodes (see \cite{moerland2020frap} as well).
\end{itemize}

\subsection{How much budget do we allocate for planning and real data collection?} \label{sec_tradeoff}
We next need to decide i) after how many real environment steps we start to plan, and ii) when we start planning for a particular state, what planning budget do we allocate? Together, these two questions determine an important trade-off in model-based RL.

\paragraph{When to start planning?}
We first need to decide how many real steps we will make before a new planning cycle. Many approaches plan after every irreversible environment step. For example, Dyna \citep{sutton1990integrated} makes up to a hundred planning steps after every real step. Other approaches collect a larger set of data before they start to plan. For example, PILCO \citep{deisenroth2011pilco} collects data in entire episodes, and replans an entire solution after a set of new real transitions has been collected. The extreme end of this spectrum is {\it batch} reinforcement learning \citep{lange2012batch}, where we only get a single batch of transition data from a running system, and we need to come up with a new policy without being able to interact with the real environment. Some methods may both start with an initial batch of data to estimate the model, but also interact with the environment afterwards \citep{watter2015embed}.

\paragraph{How much time to spend on planning?}
Once we decide to start planning, the next question is: how much planning budget do we allocate. We define a planning cycle to consist of multiple planning iterations, where each iteration is defined by fixing a new planning start state. The total planning effort is then determined by two factors: i) how many times do we fix a new start state (i.e., start a new planning iteration), and ii) how much effort does each iteration get? 

We will use Dyna \citep{sutton1990integrated} and AlphaGo Zero \citep{silver2017mastering} as illustrative examples of these two questions. In between every real environment step, Dyna samples up to a 100 one-step transitions. This means we have 100 planning iterations, each of budget 1. In contrast, in between every real step AlphaGo Zero does a single MCTS iteration, which consists of 1600 traces, each of approximate depth 200. Therefore, AlphaGo Zero performs 1 planning iteration, of budget $\sim1600*200=320.000$. The total budget per planning cycle for Dyna and AlphaGo Zero are therefore 100 and $\sim 320.000$, respectively. Note that we measure planning budget as the number of model calls here, while the true planning effort of course also depends on the computational burden of the full planning algorithm itself (which in AlphaGo Zero for example contains expensive neural network evaluations). 

Some approaches, especially the ones that target high data efficiency (see Sec. \ref{sec_dataefficiency}) in the real environment, allow for a high planning budget once they start planning. These methods for example plan until convergence on an optimal policy (given the remaining uncertainty) \citep{deisenroth2011pilco}. We call this a {\it squeezing} approach, since we attempt to squeeze as much information out of the available transition data as possible. We further discuss this approach in Sec. \ref{sec_dataefficiency}. 

\paragraph{Adaptive trade-off}
Our choice on the above two dimensions essentially specifies a trade-off between planning and real data collection, with model-free RL (no planning effort) and exhaustive search (infinite planning effort) on both extremes. Most model-based RL approaches set the above two considerations to fixed (intermediate) values. However, humans make a more adaptive trade-off \citep{keramati2011speed}, where they adaptively decide a) when to start planning, and b) how much time to spend on that plan (i.e., the two considerations discussed above). See \citet{hamrick2019analogues} for a more detailed discussion, which also incorporates more literature from human psychology. We will also return to this topic in Sec. \ref{sec_stability}.  

A few authors have investigated an adaptive trade-off between planning and acting in model-based RL. \citet{pascanu2017learning} add a small penalty for every planning step to the overall objective, which ensures that planning should provide reward benefit. This approach is very task specific. \citet{hamrick2017metacontrol} learn a meta-controller over tasks that learns to select the planning budget per timestep. In contrast to these optimization-based approaches, \citet{kalweit2017uncertainty} derive the ratio between real and planned data from the variance of the estimated Q-function. When the variance of the Q-function is high, they sample additional data from the model. This ensures that they only use ground-truth data near convergence, but accept noisier model-based data in the beginning. \citet{lu2019adaptive} propose a similar idea based on the epistemic uncertainty of the value function, by also increasing planning budgets when the uncertainty rises above a threshold. However, when we have a learned model, we probably do not want to plan too extensively in the beginning of training either (since the learned model is then almost random), so there are clear open research questions here.

\subsection{How to plan?} \label{sec_howtoplan}
The third crucial consideration is: how to actually plan? Of course, we do not aim to provide a full survey of planning methods here, and refer the reader to \citet{moerland2020frap} for a recent framework to categorize planning and RL methods. Instead, we focus on some crucial decisions we have to make for the integration, on a) the use of potential differentiability of the model, b) the direction of planning, c) the breadth and depth of the plan, and d) the way of dealing with uncertainty.  

\paragraph{Type}
One important distinction between planning methods is whether they require differentiability of the model: 

\begin{itemize}
\item {\it Discrete planning}: This is the main approach in the classic AI and reinforcement learning communities, where we make discrete back-ups which are stored in a tree, table or used as training targets to improve a value or policy function. We can in principle use any preferred planning method. Examples in the context of model-based RL include the use of probability-limited search \citep{lai2015giraffe}, breadth-limited depth-limited search \citep{franccois2019combined}, Monte Carlo search \citep{silver2008sample}, Monte Carlo Tree Search \citep{silver2017mastering,anthony2017thinking,jiang2018feedback,moerland2018a0c}, minimax-search \citep{samuel1967some,baxter1999tdleaf}, or a simple one-step search \citep{sutton1990integrated}. These methods do not require any differentiability of the model.

\item {\it Differential planning}: The gradient-based approach requires a differentiable model. If the transition and reward models are differentiable, and we specify a differentiable policy, then we can directly take the gradient of the cumulative reward objective with respect to the policy parameters. While a real world environment or simulator is by definition not differentiable, our learned model of these dynamics (for example a neural network) usually is differentiable. Therefore, model-based RL can utilize differential planning methods, exploiting the differentiability of the learned model. Note that differentiable models may also be obtained from the rules of physics, for example in differentiable physics engines \citep{degrave2019differentiable,de2018end}.

A popular example is the use of iterative linear quadratic regulator planning \citep{todorov2005generalized}, which requires a linear model, and was, for example, used as a planner in Guided Policy Search \citep{levine2013guided}. In the RL community, the gradient-based planning approach is better known as {\it value gradients} \citep{fairbank2012value,heess2015learning}. Successful examples of model-based RL that use differential planning are PILCO \citep{deisenroth2011pilco}, which differentiates through a Gaussian Process dynamics model, and Dreamer \citep{hafner2019dream} and Temporal Segment Models \citep{mishra2017prediction}, which differentiate through a (latent) neural network dynamics model.

\end{itemize}

Gradient-based planning is especially popular in the robotics and control community, since it requires relatively smooth underlying transition and reward functions. In those cases, it can be very effective. However, it is less applicable to discrete problems and sparse reward functions.


\paragraph{Direction}
We also have to decide on the direction of planning (see also Sec. \ref{sec_approximation_methods}):

\begin{itemize}

\item {\it Forward}: Forward simulation (lookahead) is the standard approach in most planning and model-based RL approaches, and actually assumed as a default in all other paragraphs of this section. We therefore do not further discuss this approach here. 

\item {\it Backward}: 
We may also learn a reverse model, which tells us which state-action pairs lead to a particular state ($s' \to s,a$). A reverse model may help spread information more quickly over the state space. This idea is better known as {\it Prioritized sweeping} (PS) \citep{moore1993prioritized}. In PS, we track which state-action value estimates have changed a lot, and then use the reverse model to identify their possible precursors, since the estimates of these state-actions are now likely to change as well. This essentially builds a search tree in the backward direction, where the planning algorithm follows the direction of largest change in value estimate.

Various papers have shown the benefit of prioritized sweeping with tabular models \citep{moore1993prioritized,dearden1999model,wiering1998efficient}, which are trivial to invert. Example that use function approximation include linear approximation \citep{sutton2012dyna}, nearest-neighbour models \citep{jong2007model}, and neural network approximation \citep{agostinelli2019solving,edwards2018forward,corneil2018efficient}. The benefit of prioritized sweeping can be large, due to its ability to quickly propagate relevant new information, but in combination with function approximation it can also be unstable. 

\end{itemize}

\begin{figure}[!t]
  \centering
      \includegraphics[width = 0.85\textwidth]{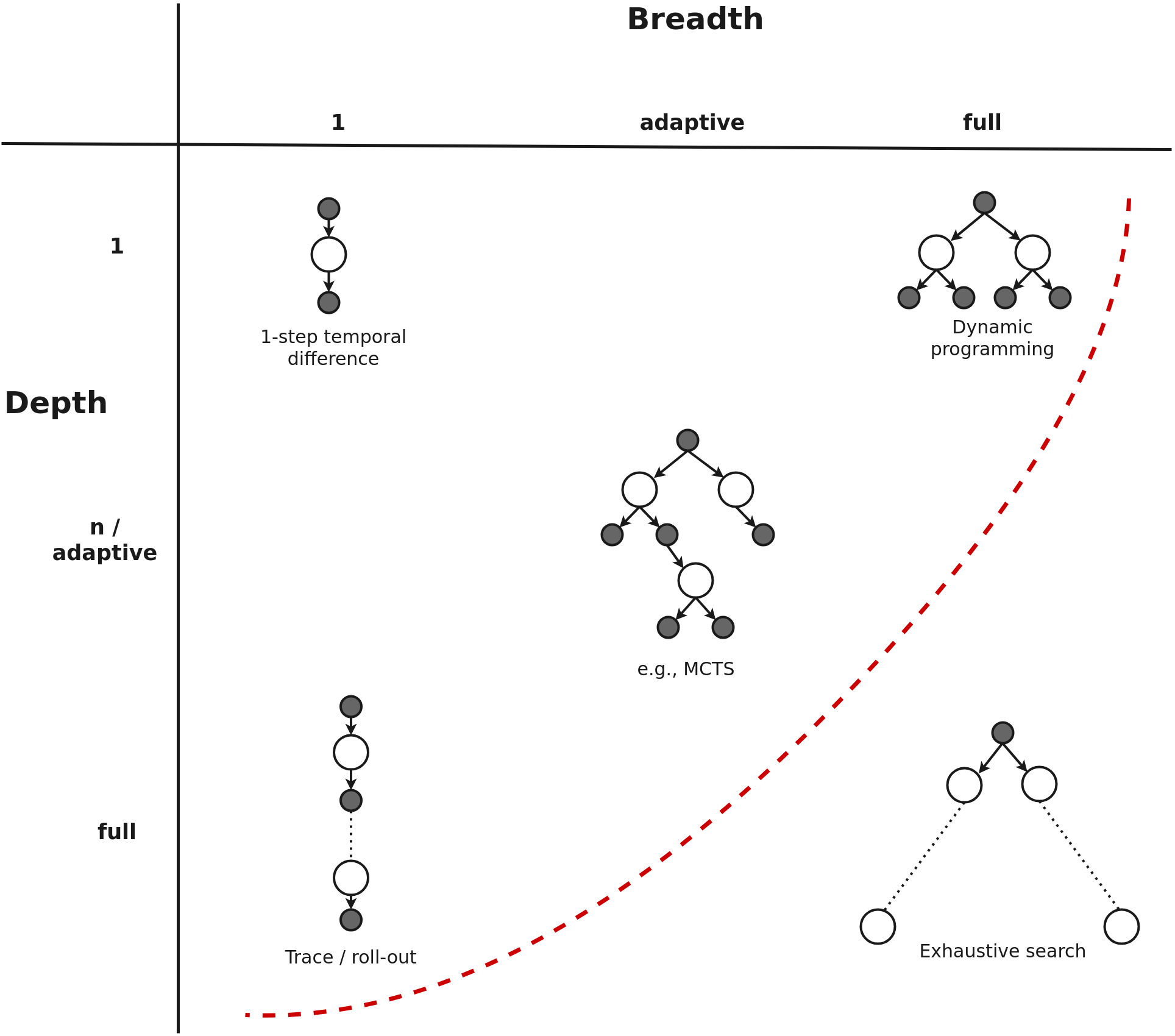}
  \caption{Breadth and depth of a single planning iteration. For every planning iteration, we need to decide on the breadth and depth of the lookahead. In practice, planning iterations usually reside somewhere left of the red dashed line, since we cannot afford a full breadth, full depth (exhaustive) search. Most planning methods, like MCTS, adaptively balance breadth and depth throughout the tree, where the breadth and depth differ throughout the tree. Figure is based on \cite{sutton2018reinforcement}, who used it to categorize different types of back-ups. A single planning iteration, which we define by fixing a new root state, can indeed be seen as a large back-up.}
    \label{fig_breadth_depth}
\end{figure}

\paragraph{Breadth and depth}
A new planning iteration starts to lookahead from a certain start state. We then still need to decide on the the breadth and the depth of the lookahead. For model-free RL approaches, breadth is not really a consideration, since we can only try a single action in a state (a breadth of one). However, a model is by definition reversible, and we are now free to choose and adaptively balance the breadth and depth of the plan. We will list the possible choices for both breadth and depth, which are summarized in Figure \ref{fig_breadth_depth}. 

For the breadth of the plan, there are three main choices:

\begin{itemize}

\item {\it Breadth = 1}: These methods only sample single transitions or individual traces from the model, and still apply model-free updates to them. Therefore, they still use a breadth of one. The cardinal example of this approach is Dyna \citep{sutton1990integrated}, which sampled additional one-step data for model-free Q-learning \citep{watkins1992q} updates. More recently, \citet{kalweit2017uncertainty} applied the above principle to deep deterministic policy gradient (DDPG) updates, \citet{kurutach2018learning} to trust region policy optimization (TRPO) updates and \citet{gu2016continuous} to normalized advantage function (NAF) updates. 

\item {\it Breadth = adaptive}: Many planning methods adaptively scale the breadth of planning. The problem is of course that we cannot afford to go full breadth and full depth, because exhaustive search is computationally infeasible. A method that adaptively scales the breadth of the search is for example Monte Carlo Tree Search \citep{kocsis2006bandit,coulom2006efficient,browne2012survey}, by means of the upper confidence bounds formula. This ensures that we do go deeper in some arms, before going full wide at the levels above. This approach was for example also used in AlphaGo Zero \citep{silver2017mastering}.

\item {\it Breadth = full}: Finally, we may of course go full wide over the action space, before we consider searching on a level deeper. This is for example the approach of Dynamic Programming, which goes full wide with a depth of one. In the context of model-based RL, few methods have taken this approach. 

\end{itemize}

For the depth of the plan, there are four choices:

\begin{itemize}

\item {\it Depth = 1}: We may of course stop after a depth one. For example, Dyna \citep{sutton1990integrated} sampled transition of breadth one and depth one. 

\item {\it Depth = intermediate}: We may also choose an intermediate search depth. RL researchers have looked at balancing the depth of the back-up for long, since it trades off bias against variance (a shallow back-up has low variance, while a deep back-up is unbiased). In the context of Dyna, \citet{holland2018effect} explicitly studied the effect of deeper roll-outs, showing that traces longer than depth 1 give better learning performance. Of course, we should be careful that deeper traces do not depart from the region where the model is accurate. 

\item {\it Depth = adaptive}: Adaptive methods for depth go together with adaptive methods for breadth. For example, an MCTS tree does not have a single depth, but usually has a different depth for many of its leafs. 

\item {\it Depth = full}: This approach samples traces in the model until an episode terminates, or until a large horizon. PILCO and Deep PILCO for example sample deep traces from their models \citep{gal2016improving}. 

\end{itemize}

This was a shallow treatment of the crucial breadth versus depth balancing in planning, which has a close relation to exploration methods as well. From a model-based RL perspective, the crucial realization is that compared to model-free RL, we can suddenly use a breadth larger than one (i.e., backtrack over multiple actions in a state). Nevertheless, many model-based RL methods still choose to stick to a breadth of one in their model samples, likely because this gives seamless integration with model-free updates. We further discuss this topic in Sec. \ref{sec_dataefficiency}.

\paragraph{Dealing with uncertainty} 
When we plan over a learned model, we usually also need to deal with the uncertainty of a learned model. There are two main approaches:

\begin{itemize}
\item {\it Data-close planning}: The first approach is to ensure that the planning iterations stay close to regions where we have actually observed data. For example, Dyna \citep{sutton1990integrated} samples start states at the location of previously visited states, and only samples one-step transitions, which ensures that we do not depart from the known region of state space. Other approaches, like Guided Policy Search \citep{levine2014learning}, explicitly constrain the new plan to be close to the current policy (which generated the data for the model). 

\item {\it Uncertainty propagation}: We may also explicitly estimate model uncertainty, which allows us to robustly plan over long horizons. Once we depart too far from the observed data, model uncertainty will increase, predictions will start to spread out over state space, and the learning signal will naturally vanish. Estimation of model uncertainty was already discussed in Sec. \ref{sec_uncertainty}. We will here focus on propagation of uncertainty over timesteps, since the next state uncertainty is of course conditioned on the uncertainty of the previous step. There are two main propagation approaches: 

\begin{itemize}
 
 \item {\it Analytic}: This propagation method fits a parametric distribution to the uncertainty at every timestep, and tries to analytically propagate the distribution over passes through the model. A well-known example is PILCO \citep{deisenroth2011pilco}, which derives closed form analytic expressions to propagate uncertainty through a Gaussian Process model. However, analytic propagation is usually not possible for more complicated models, like for example neural networks.
 
 \item {\it Sample-based}: This propagation approach, also known as {\it particle methods}, tracks the distributions of uncertainty by propagating a set of particles forward. The particles together represent the predicted distribution at a certain number of steps. Particle methods are for example used in Deep PILCO \citep{gal2016improving} and PETS \citep{chua2018deep}. Note that fitting to a distribution, or matching moments of distributions, may have a regularizing effect. Therefore, Deep PILCO \citep{gal2016improving} does propagate particles through the dynamics function, but then refits these particles to a (Gaussian) distribution at every step. See \citet{chua2018deep} for a broader discussion of uncertainty propagation approaches.
 
\end{itemize}

We may also use uncertainty to determine the {\it depth} of our value estimates. {\it Stochastic ensemble value expansion} (STEVE) \citep{buckman2018sample} reweights value targets of different depths according to their associated uncertainty, which is derived from both the value function and transition dynamics uncertainty. Thereby, we base our value estimates on those predictions which have highest confidence, which may lead to more stable learning. 

\end{itemize}

This concludes our discussion of the actual planning approach in planning-learning integration. As mentioned before, there are many more considerations in a planning algorithm, such as managing exploration (i.e., balancing breadth and depth in the planning tree). However, these challenges are not a specific aspect of planning-learning integration (but rather of RL and planning in general), and are therefore not further discussed in this survey (although we do discuss model-based exploration in Sec. \ref{sec_exploration}). 

\subsection{How to integrate planning in the learning and acting loop?} \label{sec_integration}

We have now specified how to plan (the start point, budget and planning method). However, we still need to integrate planning in the larger learning and acting loop. We now get back to Figure \ref{fig_model_based_integration}, which presented a conceptual overview of the overall training loop in planning-learning integration. We have so far focused on the planning box (arrow a), but we will now focus on the connection of planning to other aspects of the learning loop. These include: i) directing new planning iterations based on learned knowledge in value or policy functions (Fig. \ref{fig_model_based_integration}, arrow b), ii) using planning output to update learned value or policy functions (Fig. \ref{fig_model_based_integration}, arrow c), and iii) using planning output to select actions in the real world (Fig. \ref{fig_model_based_integration}, arrow d). 

\paragraph{Planning input from learned functions}
The learned value or policy functions may store much information about the current environment, which may direct the next planning iteration. We distinguish the use of value and policy information: 

\begin{itemize}
\item {\it Value priors}: The most common way to incorporate value information is through {\it bootstrapping} \citep{sutton2018reinforcement}, where we plug in the current prediction of a state or state-action value to prevent having to search deeper (reducing the depth of the search). Various model-based RL algorithms use bootstrapping in their planning approach, for example \citet{baxter1999tdleaf,silver2017mastering,jiang2018feedback,moerland2018a0c}, while is is technically also part of Dynamic Programming \citep{bellman1966dynamic}. Note that bootstrapping is also a common principle in model-free RL. We may also use the learned value function to initialize the values of the action nodes at the root of the search \citep{silver2008sample,hamrick2020combining}, which we could interpret as a form of bootstrapping at depth 0. 

\item {\it Policy priors}: We can also leverage a learned policy in a new planning iteration. Several ideas have been proposed. AlphaGo Zero \citep{silver2017mastering} uses the probability of an action as a prior multiplication term on the upper confidence bound term in MCTS planning. This gives extra exploration pressure to actions with high probability under the current policy network. Guided Policy Search (GPS) \citep{levine2013guided} penalizes a newly planned trajectory for departing too much from the trajectory generated by the current policy network. As another example, \citet{guo2014deep} let the current policy network decide at which locations to perform the next search, i.e., the policy network influences the distribution of states used as a starting point for planning (Sec. \ref{sec_startplanning}, a form of prioritization). In short, there are various ways in which we may incorporate prior knowledge from a policy into planning, although it is not clear yet which approach works best.
\end{itemize}

\paragraph{Planning update for policy or value update} 
Model-based RL methods eventually seek a global approximation of the optimal value or policy function. The planning result may be used to update this global approximation. We generally need to i) construct a training target from the search, and ii) define a loss for training. We again discuss value and policy updates separately: 

\begin{itemize}
\item {\it Value update}: A typical choice for a value target is the state(-action) value estimate at the root of the search tree. The estimate of course depends on the back-up policy, which can either be on- or off-policy. For methods that only sample a single action (i.e., use a breadth of one), like Dyna \citep{sutton1990integrated}, we may use a classic Q-learning target (one-step, off-policy). For planning cycles that do consider multiple actions in a state (that go wide and possibly deep), we can combine on- and off-policy back-ups throughout the tree in various ways. \citet{willemsen2020value} present a recent study of the different types of back-up policies in a tree search. After constructing the value target, the value approximation is usually trained on a {\it mean-squared error} (MSE) loss \citep{veness2009bootstrapping,moerland2018a0c}. However, other options are possible as well, like a cross-entropy loss between the softmax of the Q-values from the search and the Q-values of a learned neural network \citep{hamrick2020combining}.

 \item {\it Policy update}: For the policy update we again observe a variety of training targets and losses, depending on the type of planning procedure that is used. For example, AlphaGo Zero \citep{silver2017mastering} uses MCTS planning, and constructs a policy training target by normalizing the visitation counts at the root node. The policy network is then trained on a cross-entropy loss with this distribution. \citet{guo2014deep} apply the same idea with a one-hot encoding of the best action, while \citet{moerland2018a0c} cover an extension to a loss between discrete counts and a continuous policy network. As a different approach, Guided Policy Search (GPS) \citep{levine2014learning} minimizes the Kullback-Leibler (KL)-divergence between a planned trajectory and the output of the policy network. Some differential planning approaches also directly update a differentiable global representation \citep{deisenroth2011pilco}. 
 
We may also train a policy based on a value estimate. For example, Policy Gradient Search (PGS) \citep{anthony2019policy} uses the policy gradient theorem \citep{williams1992simple} to update a policy from value estimates in a tree. Note that gradient-based planning (discussed in Sec. \ref{sec_howtoplan}) also belongs here, since it directly generates gradients to update the differentiable policy.

\end{itemize}

Most of the above methods construct training targets for value or policy at the root of the search. However, we may of course also construct targets at deeper levels in the tree \citep{veness2009bootstrapping}. This extracts more information from the planning cycle. Many papers update their value or policy from both planned and real data, but other papers exclusively train their policy or value from planning \citep{ha2018recurrent,kurutach2018learning,depeweg2016learning,deisenroth2011pilco}, using real data only to train the dynamics model. 

Note that arrows b and c in Figure \ref{fig_model_based_integration} form a closed sub-loop in the overall integration. There has been much recent interest in this sub-loop, which iterates planning based on policy/value priors (arrow b), and policy/value learning based on planning output (arrow c). A successful algorithm in this class is AlphaGo Zero \citep{silver2017mastering}, which is an instance of {\it multi-step approximate real-time dynamic programming} (MSA-RTDP) \citep{efroni2019multi}. MSA-RTDP extends the classic DP ideas by using a `multi-step' lookahead, learning the value or policy (`approximate'), and operating on traces through the environment (`real-time'). \citet{efroni2019multi} theoretically study MSA-RTDP, showing that higher planning depth $d$ decreases sample complexity in the real environment at the expense of increased computational complexity. Although this is an intuitive result, it does prove that planning may lead to better informed real-world decisions, at the expense of increased (model-based) thinking time. In addition, iterated planning and learning may also lead to more stable learning, which we discuss in Sec. \ref{sec_stability}. 

\paragraph{Planning output for action selection in the real environment}
We may also use planning to select actions in the real environment. While model-free RL has to use the value or policy approximation to select new action in the environment (Fig. \ref{fig_model_based_integration}, arrow e), model-based RL may also select actions directly from the planning output (Fig. \ref{fig_model_based_integration}, arrow d). Some methods only use planning for action selection, not for value/policy updating \citep{tesauro1997online,silver2008sample}, for example because planning updates can have uncertainty. However, many methods actually combine both uses \citep{silver2017mastering,silver2018general,anthony2017thinking,moerland2018a0c}. 

Selection of the real-world actions may happen in a variety of ways. First of all, we may greedily select the best action from the plan. This is the typical approach of methods that `plan over a learned model' (Table \ref{tab_model_based_boundaries2}). The cardinal example in this group are {\it model predictive control} (MPC) or {\it receding horizon control} approaches. In MPC, we find the greedy action of a $k$-step lookahead search, execute the greedy action, observe the true next state, and repeat the same procedure from there. The actual planning algorithm in MPC may vary, with examples including iLQR \citep{watter2015embed},  direct optimal control \citep{nagabandi2018neural,chua2018deep}, Dijkstra's algorithm \citep{kurutach2018learning}, or repeated application of an inverse model \citep{agrawal2016learning}. MPC is robust to (changing) constraints on the state and action space \citep{kamthe2017data}, and is especially popular in robotics and control tasks. 

Note that we do not have to execute the greedy action after a planning cycle ends, but can introduce additional exploration noise (like Dirichilet noise in \citet{silver2017mastering}). We can also explicitly incorporate exploration criteria in the planning process, which we may call 'plan to explore' \citep{sekar2020planning,lowrey2018plan}. For example, \citet{dearden1998bayesian} explore based on the value of perfect information' (VPI), which estimates from the model which action has the highest potential to change the greedy policy. Indeed, planning may identify action sequences that perform `deep exploration' towards new reward regions, which one-step exploration methods would fail to identify due to jittering behaviour \citep{osband2016deep}.

This concludes our discussion of the main considerations in planning-learning integration. Table \ref{table_framework4} summarizes the framework, showing the potential decisions on each dimension. 

 \begin{figure}[!p]
  \centering
      \includegraphics[width = 0.75\textwidth]{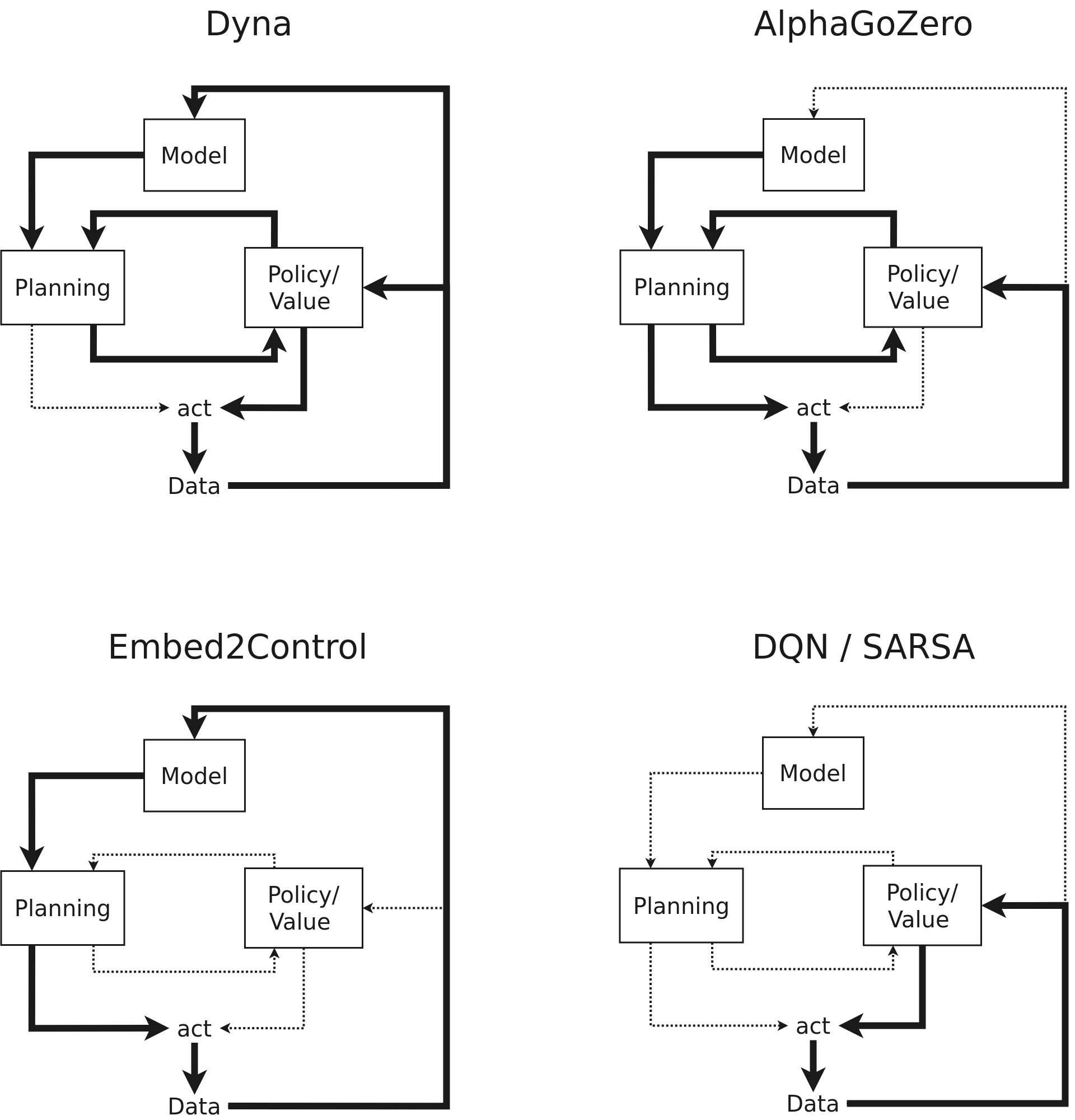}
  \caption{Comparison of planning and learning algorithms, based on the general visualization of learning/planning integration from Figure \ref{fig_model_based_integration}. Thick lines are used by an algorithm. Dyna \citep{sutton1991dyna} (top-left) is an example of model-based RL with a learned model. AlphaGo Zero \citep{silver2017mastering} (top-right) is an example of model-based RL with a known model. Note that therefore the model does not need updating from data. Embed2Control \citep{watter2015embed} (bottom-left) is an example of planning over a learned model. For comparison, the bottom right shows a model-free RL algorithm, like Deep Q-Network \citep{mnih2015human} or SARSA \citep{rummery1994line} with eligibility traces}.
    \label{fig_model_based_comparisons}
\end{figure} 



\begin{sidewaystable}[!p]
\footnotesize
\caption{ \footnotesize Systematic comparison of different model-based RL algorithms on the dimensions of planning-learning integration (Sec. \ref{sec_model_using}). Colour coding: green = model-based RL with a learned model, red = model-based RL with a known model, blue = planning over a learned model (see Table \ref{tab_model_based_boundaries2}). Uncertainty estimation methods: GP = Gaussian Process, BE = bootstrap ensemble. Uncertainty propagation methods: Par = parametric propagation, Sam = sample-based propagation (particle methods). $\dagger$ = Before learning, the authors collect an initial batch of training data for the model. The number of real steps before the {\it first} plan is therefore 3.000-30.000, depending on the task. Afterwards, they start to interact with the environment, planning at every step. $\star$ = gradient-based planners improve a reference trajectory based on gradients. Although there is only one trajectory, the gradient does implicitly go wide over the actions, since it tells us in which direction the continuous action should be moved.} \label{table_overview4}

\centering
\begin{adjustbox}{width=\textheight}
\begin{tabular}{p{3cm} P{1.3cm} P{1.4cm} P{1.7cm} P{1.3cm} P{1.3cm} P{1.3cm} P{1.7cm} P{1.4cm} P{1.3cm} P{1.3cm}}
\toprule

{\bf Paper}  &  {\bf Start state} & \multicolumn{2}{c}{\bf Budget}  & \multicolumn{4}{c}{\bf How to plan?} & \multicolumn{3}{c}{\bf Integration within learning loop}  \\

\cline{3-4} \cline{5-8} \cline{9-11} \\

 &   & Real steps before plan & Budget per planning cycle & Type & Direction &  Breadth \& depth  & Uncertainty & Input from value/policy & Output to value/policy & Output for action selection \\

\midrule
\cellcolor{green!25} Dyna \citep{sutton1990integrated}\newline & Visited &   1 & 10-100 steps & Discrete & Forward &   B=1, D=1 & Data-close &   V & V &    - \\

\cellcolor{green!25} Prioritized sweeping \citep{moore1993prioritized} & Prioritized &   1 & 10 steps & Discrete &  Backward &    B=full, D=1 &    - &   V  & V &   - \\

\cellcolor{green!25} PILCO \citep{deisenroth2011pilco} & Start & Episode &   $\uparrow$ \newline (convergence) & Gradient & Forward & B$>$1$^\star$, D=full &   Uncertainty \newline (GP + Par) &   P & P &   - \\

\cellcolor{green!25} Guided policy search \citep{levine2014learning} & Current & Episode & 5-20 rollouts & Gradient  & Forward & B=5-40, D=full  & Data-close & P & P & - \\

\cellcolor{red!25} AlphaGo Zero \citep{silver2017mastering} & Current &   1 &   $\sim$320.000 & Discrete & Forward &   adaptive &   -  &    V+P & P &    $\checkmark$ \\

\cellcolor{red!25} SAVE \citep{hamrick2020combining} & Current & 1 & 10-20 & Discrete & Forward & adaptive &  -  & V & V & $\checkmark$  \\

\cellcolor{blue!25} Embed2Control \citep{watter2015embed} & Current & 1$^\dagger$ & MPC depth 10 & Gradient & Forward &   B$>$1$^\star$, D=10 & - & - & - & $\checkmark$ \\

\cellcolor{blue!25} PETS \citep{chua2018deep} & Current & 1 & MPC depth 10-100 & Discrete & Forward & B$>$1, D=10-100 & Uncertainty \newline (BE + Sam)  & P & - & $\checkmark$ \\

\bottomrule 
\end{tabular} 
\end{adjustbox}
\end{sidewaystable}

\subsection{Conceptual comparison of approaches}
This section discussed the various ways in which planning and learning can be integrated. We will present two summaries of the discussed material. First of all, Figure \ref{fig_model_based_comparisons} summarizes the different types of connectivity that may be present in planning-learning integration. The figure is based on the scheme of Figure \ref{fig_model_based_integration}, as used throughout this section, and the classification of model-based RL methods described in Table \ref{tab_model_based_boundaries2}.  

We see how well-known model-based RL algorithms like Dyna \citep{sutton1991dyna} and AlphaGo Zero \citep{silver2017mastering} use different connectivity. For example, Dyna learns a model, which AlphaGo Zero assumes to be known, and AlphaGo Zero selects actions from planning, while Dyna uses the learned value approximation. The bottom row shows Embed2Control \citep{watter2015embed}, a method that only plans over a learned model, and completely bypasses any global policy or value approximation. For comparison, the bottom-right of the figure shows a model-free RL approach, like DQN \citep{mnih2015human} or SARSA \citep{rummery1994line} with eligibility traces. 

As a second illustration, Table \ref{table_overview4} compares several well-known model-based RL algorithms on the dimensions of our framework for planning-learning integration (Table \ref{table_framework4}). We see how different integration approaches make different choices on each of the dimensions. It is hard to judge whether some integration approaches are better than others, since they are generally evaluated on different types of tasks (more on benchmarking in Sec. \ref{sec_discussion}. The preferred choices of course also depend on the specific problem and available resources. For example, a problem like Go requires are relatively high planning budget per timestep \citep{silver2017mastering}, while for smaller problems a lower planning budget per timestep may suffice. Gradient-based planning can be useful, but is mostly applicable to continuous control tasks, due to the relatively smooth dynamics. For many considerations, there are both pros and cons. Usually, the eventual decisions depends on hyperparameter optimization and the type of benefit (of model-based RL) we aim for, which we will discuss in Sec \ref{sec_benefits}. 


\section{Implicit Model-based Reinforcement Learning} \label{sec_implicit_mbrl}
We have so far discussed the two key steps of model-based RL: 1) model learning and 2) planning over the model to recommend an action or improve a learned policy or value function. All the methodology discussed so far was {\it explicit}, in a sense that we manually specified each part of the algorithm. This is the classical, explicit approach to model-based RL (and to algorithm design in general), in which we manually design the individual elements of the algorithms. 

\begin{figure}[p]
  \centering
      \includegraphics[width = 1.0\textwidth]{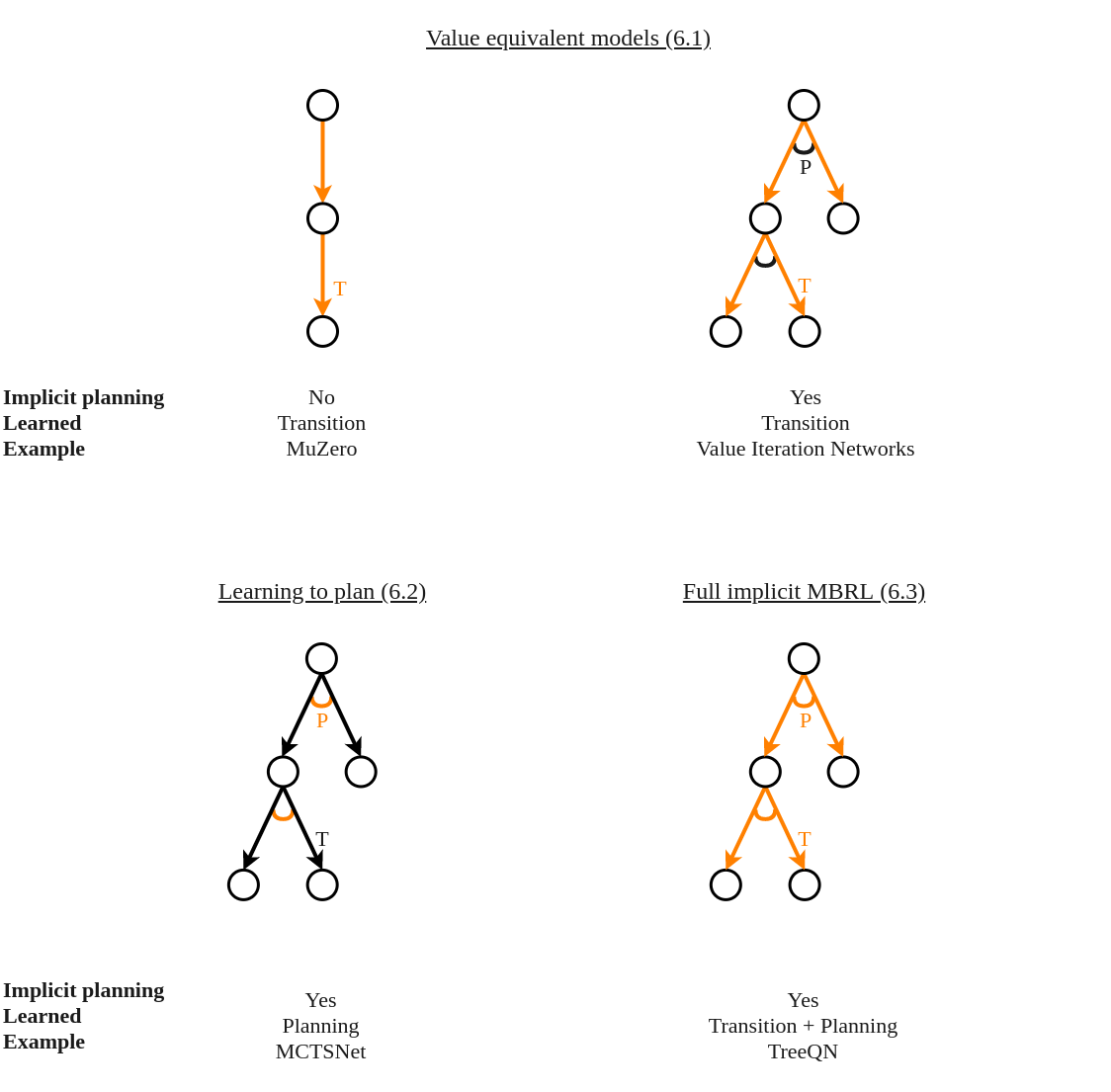}
  \caption{Categories of implicit model-based RL (Sec. \ref{sec_implicit_mbrl}). Graphs schematically illustrate a differentiable computation graph, with transitions (T) denoted as lines, and planning/policy improvement (P) denoted as an arc between alternative actions (which are not explicitly drawn). The depiction of planning (policy improvement in the graph) is of course conceptual, and may in practice involve several networks (e.g., an action selection network and a back-up network). For each graph, black lines are known (in differentiable form), while orange lines are learned. Top-left: Value equivalent model in the policy evaluation setting, of which MuZero \citep{schrittwieser2019mastering} is an example. Top-right: Value equivalent model with implicit planning, of which Value Iteration Networks \citep{tamar2016value} are an example. Bottom-left: Learning to plan, of which MCTSNet \citep{guez2018learning} is an example. Bottom-right: Full implicit model-based RL, of which TreeQN \citep{farquhar2018treeqn} is an example.}
    \label{fig_implicit_model_based_RL}
\end{figure} 

An interesting observation about the above process is that, although we may manually design various aspects of model-based RL algorithms, we ultimately only care about one thing: identifying the (optimal) value or policy. In other words, the entire model-based RL procedure (model learning,  planning, and possibly integration in value/policy approximation) can from the outside be seen as a single optimization objective, since we want it to predict an (optimal) action or value. This intuition leads us to the field of {\it implicit} model-based RL. The common idea underneath all these approaches is to take one or more aspects of the model-based RL process and optimize these for the ultimate objective, i.e., (optimal) value or policy computation. 

In particular, we will focus on methods that use gradient-based optimization. In those cases, we embed (parts of) the model-based RL process within a differentiable computational graph, which eventually outputs a value or action recommendation. Since the graph remains end-to-end differentiable, we may optimize one or more elements of our model-based RL procedure for an eventual value prediction or action recommendation. One would be tempted to therefore call the field `end-to-end model-based RL', but note that the underlying principles are more general, and could also work with gradient-free optimization. 

We may use implicit model-based RL to replace each (or both) of the steps of explicit model-based RL: 1) to optimize a transition model and 2) to optimize for the actually planning procedure (i.e., some form of policy optimization). This leads to the possible settings shown in Fig. \ref{fig_implicit_model_based_RL}. The figure schematically shows the different types of computational graphs we may construct, and which elements of this graph are optimized for (shown in orange). We will discuss these possible settings shown in the later part of this section. 

The differentiable computational graphs of course need to be optimized against some outer objective, for which there are two options. First, we may train the graph for its ability to predict the correct (optimal) value (an {\it RL loss}). This value is frequently obtained from a standard model-free RL target constructed from observed traces. The second option is to train the graph against its ability to output the correct (optimal) action or policy (an {\it imitation loss}). Such knowledge may either be available from expert demonstrations, or can be obtained from running a separate model-free RL agent. The underling intuition is to first optimize the model and/or planning procedure against correct value or action targets in a task, which may afterwards lead to superior performance in the same task, or generalization to other tasks. 

The remainder of this section will discuss the three possible forms of computational graphs shows in Fig. \ref{fig_implicit_model_based_RL}: value equivalent models (Sec. \ref{sec_value_equivalent}), where we only optimize the transition dynamics in the graph, learning to plan (Sec. \ref{sec_learning_to_plan}), where we only optimize the planning operations, and full implicit model-based RL (Sec. \ref{sec_combination_implicit}), where we jointly optimize the transition model and planning operations. A structured overview of the papers we discuss is provided in Table \ref{table_implicit_mbrl}. 

\subsection{Value equivalent models} \label{sec_value_equivalent}
Standard model learning approaches, as discussed in Section \ref{sec_model_learning}, learn a forward model that predicts the next state of the environment. However, such models may predict several aspects of the state that are not relevant for the value. In some domains, the forward dynamics might be complicated to learn, but the aspects of the dynamics that are relevant for value prediction might be much simpler. \citet{grimm2020value} theoretically study this idea, which they name {\it value equivalent models}. Value equivalent models are unrolled inside the computation graph to predict a future value (or action), instead of a future state. As such, these models are enforced to emphasize value-relevant characteristics of the environment.  

An example of a successful value-equivalent approach is MuZero \citep{schrittwieser2019mastering}. To learn a transition function, MuZero internally unrolls a model to predict a multi-step, action-conditional value (Fig. \ref{fig_implicit_model_based_RL}, top-left). The graph is then optimized for its ability to predict a model-free value estimate at each step. The obtained value-equivalent model is subsequently use in an MCTS procedure, which achieved state-of-the-art performance in the Chess, Go and Shogi. Other successful examples of this approach are Value Prediction Networks (VPN) \citep{oh2017value} and the Predictron \citep{silver2017predictron}. 

A crucial aspect of the above methods is that they unroll a single trace, and therefore train in the policy evaluation setting. In contrast, we may also optimize the transition model in an implicit planning graph, which does contain policy improvement (Fig. \ref{fig_implicit_model_based_RL}, top-right). Two example papers that take this approach are Value Iteration Networks (VIN) \citep{tamar2016value} and Universal Planning Networks (UPN) \citep{srinivas2018universal}. Both papers specify a known, differentiable planning procedure in the graph (VIN embeds value iteration, UPN embeds value-gradient planning). The entire planning procedure consist of multiple cycles through the transition model and planner, which is then optimized for its ability to predict a correct optimal action or value. Since VINs and UPNs do incorporate policy improvement in the computational graph, they may learn slightly different aspects of the dynamics than MuZero and VPN (i.e., VIN/UPN will only emphasize aspects necessary to predict the optimal action/value, while MuZero/VPN will emphasize aspects necessary to make correct multi-step predictions for all observed action sequences). 

\begin{sidewaystable}[p]
\scriptsize
\caption{\footnotesize Comparison of implicit model-based RL approaches. {\bf Rows}: algorithm colour coding, yellow = explicit model-based RL (for comparison), green = value equivalent models (Sec. \ref{sec_value_equivalent}), red = learning to plan (Sec. \ref{sec_learning_to_plan}), blue = combination of value equivalent models and learning to plan (Sec. \ref{sec_combination_implicit}). {\bf Columns}: Implicit planning implies some form of policy improvement in the computation graph. Learning to plan implies that this improvement operation is actually optimized. For planning, we shortly mention the specific planning structure between brackets. MCTS = Monte Carlo Tree Search, iLQR = iterative Linear Quadratic Regulator, RNN = recurrent neural network.} \label{table_implicit_mbrl}

\centering
\begin{adjustbox}{width=\textheight}
\begin{tabular}{ p{4.5cm}  P{2.0cm}  P{2.0cm} P{2.0cm} P{2.0cm}  P{2.0cm} P{2.0cm} }
\toprule
\multicolumn{1}{c}{\bf Paper}   & \multicolumn{3}{|c|}{\bf Model} & \multicolumn{3}{c}{\bf Planning}  \\

  & Known state-prediction & Learned state-prediction &  Learned value-equivalent & Explicit & Implicit & Learning to plan \\  
  
\bottomrule
\cellcolor{yellow!25} AlphaZero \citep{silver2018general} \newline & x  &  & & x (MCTS) & &  \\
\cellcolor{yellow!25} Dyna \citep{sutton1990integrated} \newline &  & x  & & x (one-step) & &    \\
\cellcolor{yellow!25} Embed to Control (E2C) \citep{watter2015embed} &  & x  & & x (iLQR) &  & \\

\cellcolor{green!25} MuZero \citep{schrittwieser2019mastering} \newline &  &  & x   & x (MCTS) &  & \\
\cellcolor{green!25} Value prediction networks (VPN) \citep{oh2017value} &  &  & x & x ($b$-best, depth-$d$ plan) &  & \\
\cellcolor{green!25} Predictron \citep{silver2017predictron} \newline  &  &  & x  & x (Roll-out) &  & \\

\cellcolor{green!25} Value Iteration Networks (VIN) \citep{tamar2016value} &  &  & x &  & x (value iteration)  & \\

\cellcolor{green!25} Universal Planning Networks (UPN) \citep{srinivas2018universal} &  &  & x &  & x (grad-based planning) & \\

\cellcolor{red!25} MCTSNet \citep{guez2018learning} \newline & x &  &  & & x & x (MCTS aggregation) \\

\cellcolor{red!25} Imagination-augmented agents (I2A) \citep{racaniere2017imagination} &  & x  & & & x & x (Roll-out aggregation) \\

\cellcolor{red!25} Imagination-based planner (IBP) \citep{pascanu2017learning} &  & x  &  & & x & x (Tree constr. + aggregation) \\

\cellcolor{blue!25} TreeQN \citep{farquhar2018treeqn} \newline &  &  & x &  & x & x (Tree aggregation) \\

\cellcolor{blue!25} Deep Repeated ConvLSTM (DRC) \citep{guez2019investigation} &  &  & x & & x & x (RNN) \\

\bottomrule
\end{tabular}
\end{adjustbox}
\end{sidewaystable}

\subsection{Learning to plan} \label{sec_learning_to_plan}
We may also use the implicit planning idea to optimize the planning operations themselves (Fig. \ref{fig_implicit_model_based_RL}, bottom-left). So far, we encountered two ways in which learning may enter model-based RL: i) to learn a dynamics model (Sec. \ref{sec_model_learning}), and ii) to learn a value or policy function (from planning output) (Sec. \ref{sec_model_using}). We now encounter a third level in which learning may enter model-based RL: to learn the actual planning operations (and its integration with a learned value or policy function). End-to-end learning has of course shown success in other machine learning fields, such as computer vision, where it has gradually replaced manually constructed features. A similar trend starts to appear for entire algorithms, which we may better learn through optimization than manually specify. This idea is known as {\it algorithmic function approximation} \citep{guez2019investigation}, where our learned approximator makes multiple internal cycles (like a recurrent neural network) to improve the quality of its prediction. Note that this differs from standard RNNs, where the recurrence is typically used to deal with additional inputs or outputs (e.g., in the time dimension), while in algorithmic function approximation the {\it additional internal cycles improve the quality of the predictions}.

An example of learning to plan are MCTSNets \citep{guez2018learning}. This approach specifies elements of the MCTS algorithm, like selection, back-up and final recommendation, as neural networks, and optimizes these against the ability to predict the correct optimal action in the game Sokoban. While MCTSNets assume a known dynamics model, Imagination-augmented agents (I2A) \citep{racaniere2017imagination} first separately learn a differentiable dynamics model (in the standard way), and then learn how to aggregate information from roll-outs in this model through the implicit-model-based RL approach (which shows we can actually combine conventional and implicit model-based RL approaches). While both these algorithms (MCTSNets and I2A) still include some manual design in their planning procedure (like the order of node expansion), Imagination-based planner (IBP) \citep{pascanu2017learning} also includes a differentiable manager network that in each iteration decides whether we continue planning and from which state we will plan. This gives the algorithm almost full freedom in the algorithmic planning space, and the authors therefore also include a cost for simulation, which ensures that the agent will not continue to plan forever. Results show that the agent indeed learns both how to plan and for how long to plan.  

\subsection{Combined learning of models and planning} \label{sec_combination_implicit}
We may also combine both ideas introduced in the previous sections (value equivalent models and learning to plan): if we specify a parameterized differentiable model and a parameterized differentiable planning procedure, then we can optimize the resulting computational graph jointly for the model and the planning operations (Fig. \ref{fig_implicit_model_based_RL}, bottom-right). This of course creates a harder optimization problem, since the gradients for the planner depend on the quality of the model, and vice versa. However, it is the most end-to-end approach to model-based RL we can imagine, as all aspects discussed in Sections \ref{sec_model_learning} and \ref{sec_model_using} get wrapped into a single optimization. 

An example approach in this category is TreeQN \citep{farquhar2018treeqn}. From the outside, TreeQN looks like a standard value network, but internally it is structured like a planner. The planning algorithm of TreeQN unrolls itself up to depth $d$ in all directions, and aggregates the output of these predictions through a back-up network, which outputs the value estimate for the input state. We then optimize both the dynamics model and the planning procedure (back-up network) against a standard RL loss. While TreeQN still has some internal structure, full algorithmic freedom is provided by the Deep Repeated \mbox{ConvLSTM} (DRC) \citep{guez2019investigation}. DRC is a high-capacity recurrent neural network without any planning or MDP specific internal structure. It is therefore entirely up to the DRC to approximate both an appropriate (value-equivalent) model and an appropriate planning procedure, which the authors call {\it model-free planning}. Indeed, DRC does show characteristics of planning after training, like an increase in test performance with increasing computational budget.

\vspace{0.5cm}

An overview of the discussed papers in this section is provided in Table \ref{table_implicit_mbrl}. The strength of the implicit model-based RL approach is tied to the strength of optimization in general, which, as other fields of machine learning have shown, may outperform manual design. Moreover, value equivalent models may be beneficial in tasks where the dynamics are complicated, but the dynamics relevant for value estimation are easier to learn. On the other hand, implicit model-based RL has its challenges as well. For value-equivalent transition models, all learned predictions focus on the value and reward information, which is derived from a scalar signal. These methods will therefore likely not capture all relevant aspects of the environment, which may be problematic for transfer. A similar problem occurs for learning to plan, where we risk that our planner will learn to exploit task-specific characteristics, which does not generalize to other tasks. The true solution to these problems is of course to train on a wide variety of tasks, which is computationally demanding, while implicit model-based RL is already computationally demanding in itself (the computational graphs grow large, and the optimization can be unstable). Model-based RL therefore faces the same fundamental question as many other artificial intelligence and machine learning directions: to what extend should our systems incorporate human priors ({\it explicit}), or rely on black-box optimization instead ({\it implicit}).

\section{Benefits of Model-based Reinforcement Learning} \label{sec_benefits}
Model-based RL may provide several benefits, which we will discuss in this section. However, in order to identify benefits, we first need to discuss performance criteria, and establish terminology about the two types of exploration in model-based RL.  

\paragraph{Performance criteria}
There are two main evaluation criteria for (model-based) RL algorithms:
\begin{itemize}
\item {\it Cumulative reward/optimality}: the quality of the solution, measured by the expected cumulative reward that the solution achieves. 
\item {\it Time complexity}: the amount of time needed to arrive at the solution, which actually has three subcategories:
 \begin{itemize}
\item Real-world sample complexity: how many unique trials in the real (irreversible) environment do we use? 
\item Model sample complexity: how many unique calls to a (learned) model do we use? This is an infrequently reported measure, but may be a useful intermediate.
\item Computational complexity: how much unique operations (flops) does the algorithm require. 
\end{itemize}

\end{itemize}

Papers usually report learning curves, which show optimality (cumulative return) on the y-axis and one of the above time complexity measures on the x-axis. As we will see, model-based RL may actually be used to improve both measures.

\begin{figure}[t]
  \centering
      \includegraphics[width = 1.0\textwidth]{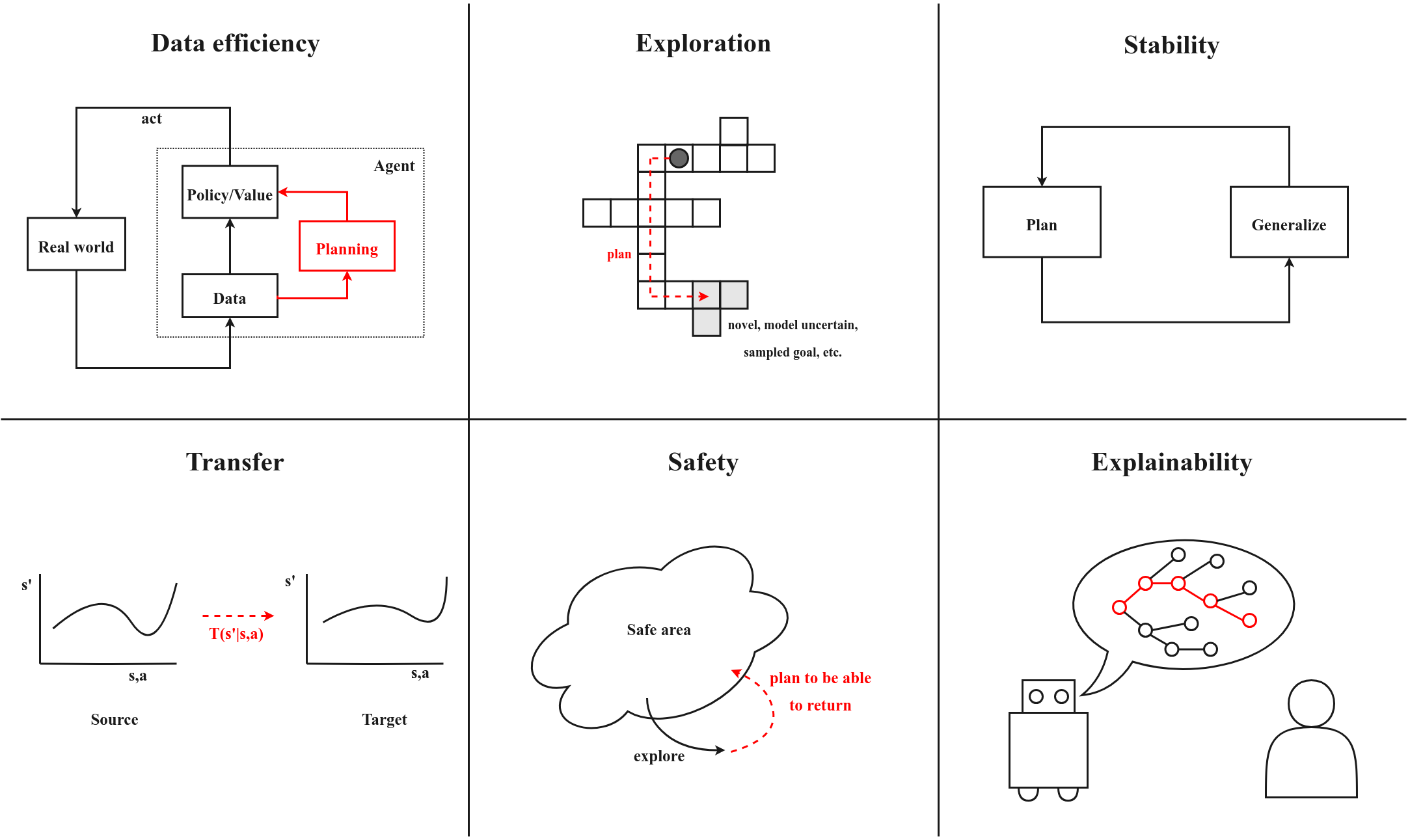}
  \caption{Benefits of model-based reinforcement learning, as discussed in Section \ref{sec_benefits}.}
    \label{fig_modelrl_benefits}
\end{figure}

We will now discuss the potential benefits of model-based RL (Figure \ref{fig_modelrl_benefits}). First, we will discuss enhanced data efficiency (Sec. \ref{sec_dataefficiency}), which uses planning (increased model sample complexity) to reduce the real-world sample complexity. Second, we discuss exploration methods that use model characteristics (Sec. \ref{sec_exploration}). As a third benefit, we discuss the potential of model-based RL with a known model to reach higher asymptotic performance (optimality/cumulative reward) (Sec. \ref{sec_stability}). A fourth potential benefit is transfer (Sec. \ref{sec_transfer}), which attempts to reduce the sample complexity on a sequence of tasks by exploiting commonalities. Finally, we also shortly touch upon safety (Sec. \ref{sec_safety}), and explainability (Sec. \ref{sec_explainability}).  

\subsection{Data Efficiency} \label{sec_dataefficiency}
A first approach to model-based RL uses planning to reduce the real-world sample complexity. Real-world samples are expensive, both due to wall-clock time restrictions and hardware vulnerability. Enhanced data efficiency papers mostly differ by how much effort they invest per planning cycle (Sec. \ref{sec_tradeoff}). A first group of approaches tries to {\it squeeze} out as much information as possible in every planning loop. These typically aim for maximal data efficiency, and apply each planning cycle until some convergence criterion. Note that batch reinforcement learning \citep{lange2012batch}, where we only get a single batch of data from a running system and need to come up with an improved policy, also falls into this group. The second group of approaches continuously plans in the background, but does not aim to squeeze all information out of the current model. 

\begin{itemize}
\item {\it Squeezing}: The squeezing approach, that plans from the current state or start state until (near) convergence, has theoretical motivation in the work on Bayes-adaptive exploration \citep{duff2002optimal,guez2012efficient}. All data efficiency approaches crucially need to deal with model uncertainty, which may be estimated with a Bayesian approach \citep{guez2012efficient,asmuth2009bayesian,castro2007using}. These approaches are theoretically optimal in real world sample complexity, but do so at the expense of high computational complexity, and crucially rely on correct Bayesian inference. Due to these last two challenges, Bayes-adaptive exploration is not straightforward to apply in high-dimensional problems. 

Many empirical papers have taken the squeezing approach, at least dating back to \citet{atkeson1997comparison} and \citet{boone1997efficient}. We will provide a few illustrative examples. A breakthrough approach was PILCO \citep{deisenroth2011pilco}, which used Gaussian Processes to account for model uncertainty, and solved a real-world Cartpole problem in less than 20 seconds of experience. PETS \citep{chua2018deep} used a bootstrap ensemble to account for uncertainty, and scales up to a 7 degrees-of-freedom (DOF) action space, while model-based policy optimization (MBPO) \citep{janner2019trust}, using a similar bootstrap ensemble for model estimation, even scales up to a 22 DOF humanoid robot (in simulation). Embed2Control \citep{wahlstrom2015pixels} managed to scale model-based RL to a pixel input problem. Operating on a 51x51 pixel view of Pendulum swing-up, they show a 90\% success rate after 15 trials of a 1000 frames each. 

\item {\it Mixing}: The second group of approaches simply mixes model-based updates with model-free updates, usually by making model-based updates (in the background) throughout the (reachable) state space. The original idea dates back to the Dyna architecture of \citet{sutton1990integrated}, who reached improved data efficiency of up to 20-40x in a gridworld problem. In the context of high-dimensional function approximation, \citet{gu2016continuous,nagabandi2018neural} used the same principle to reach a rough 2-5 times improvement in data efficiency. 

An added motivation for the mixing approach is that we may still make model-free updates as well. Model-free RL generally has better asymptotic performance than model-based RL with a learned model. By combining model-based an model-free updates, we may speed-up learning with the model-based part, while still reaching the eventual high asymptotic performance of model-free updates. Note that model-based RL with a known model may actually reach higher asymptotic performance (Sec. \ref{sec_stability}) than model-free RL, which shows that the instability is really caused by the uncertainty of a learned model.
\end{itemize}

In short, model-based RL has a strong potential to increase data efficiency, by means of two-phase exploration. Strong improvements in data efficiency have been shown, but are not numerous, possibly due to the lack of stable uncertainty estimation in high-dimensional models, or the extensive amount of hyperparameter tuning required in these approaches. Nevertheless, good data efficiency is crucial for scaling RL to real world problems, like robotics \citep{kober2013reinforcement}, and is a major motivation for the model-based approach. 

\subsection{Exploration} \label{sec_exploration}
The trade-off between exploration and exploitation is a crucial topic within reinforcement learning, in which model-based approaches can play an important role. There are two important considerations that determine whether a particular exploration approach is model-based, as visualized in Table \ref{table_exploration_methods}. First, we need to distinguish {\it one-phase versus two-phase exploration} (Table \ref{table_exploration_methods}, columns). Model-free RL methods and pure planning methods use `one-phase exploration': they use the same exploration principle in the entire algorithm, i.e., either within a trace (model-free RL) or within a planning tree. In contrast, model-based RL agents use `two-phase exploration', since they may combine 1) an exploration strategy within the planning cycle, and 2) a (usually more conservative/greedy) strategy for the irreversible (real environment) step. In the case of model-based RL with a learned model, the aim of this approach is usually to reduce real world sample complexity at the expense of increased model sample complexity. This has a close relation to the previous section (on data efficiency), although we there mostly focused on additional model-based back-ups, not exploration. In the case of model-based RL with a known model we also observe two-phase exploration, like confidence bound methods inside the tree search and Dirichlet noise for the real steps in AlphaGo Zero \citep{silver2017mastering}. However, with a known model (in which case planning and real steps both happen in the same reversible model), the second phase rather seems a pruning technique, to ensure that we stop planning at some point and advance to a next state. 

The second important distinction is between {\it value-based and state-based exploration (intrinsic motivation)} (Table \ref{table_exploration_methods}, rows). Value-based methods based their exploration strategy on the current value estimates of the available actions. Actions with a higher value estimate will also get a higher probability of selection, where the perturbation may for example be random \citep{plappert2017parameter,mnih2015human} or based on uncertainty estimates around these values \citep{auer2002using,osband2016deep,moerland2017efficient}. The model-based alternative is to use `state-based' exploration. In this case, we do not determine the exploration potential of a state based on reward or value relevancy, but rather based on state-specific, reward independent properties derived from the interaction history with that state. A state may for example be interesting because it is novel or has high uncertainty in its model estimates. These approaches are better known as {\it intrinsic motivation} (IM) \citep{chentanez2005intrinsically}. 

The two above distinctions together lead to four possible combinations, as visualized in the cells of Table \ref{table_exploration_methods}. We define {\it model-based exploration} as `any exploration approach that uses either state-based exploration and/or two-phase exploration' (indicated by the grey boxes in Table \ref{table_exploration_methods}). We therefore consider all state-based exploration methods as model-based RL. State-based exploration methods often use model-based characteristics or a density model over state space (which in the tabular setting can directly be derived from a tabular model), and therefore have a close relation to model-based RL, even when it is applied without actual planning (one-phase).

\begin{table}[t]
\centering
\footnotesize
\caption{Categories of exploration methods. Grey cells are considered `model-based exploration', since they either use state-based characteristics and/or plan over the model to find better exploration decisions (two-phase exploration).}
\label{table_exploration_methods}
\begin{adjustbox}{width=\textwidth}
\begin{tabular}{ p{4.0cm} P{4.0cm} P{4.0cm}}
\toprule
   & \bfseries{One-phase exploration} & \bf Two-phase exploration \\
   \hline

\bf Value-based exploration & e.g., $\epsilon$-greedy on value function & \cellcolor{gray!20} e.g., planning to find a high value/reward region \\ 
\bfseries{State-based exploration} & \cellcolor{gray!20} e.g., intrinsic reward for novelty without planning & \cellcolor{gray!20} e.g., planning towards an novel (goal) state \\ 
 
\bottomrule
\end{tabular}
\end{adjustbox}
\end{table}

\paragraph{Considerations in exploration}
To get a better understanding of the main challenges in (model-based) exploration, we will first discuss the most important general challenges in exploration: 

\begin{itemize}
\item {\it Shallow versus deep exploration}: Every exploration method can be classified as either shallow or deep. {\it Shallow} exploration methods redecide on their exploratory decision at every timestep. In the model-free RL context, $\epsilon$-greedy exploration is a good example of this approach. The potential problem of these approaches is that they do not stick with an exploratory plan over multiple timestep. This may lead to `jittering' behaviour, where we make an exploratory decision in a state, but decide to undo it at the next timestep. 

Intuitively, we rather want to fix an interesting exploration target in the future, and commit to a sequence of actions to actually get there. This approach is known as {\it deep} exploration \citep{osband2016deep} (note that `deep' in this case has nothing to do with the depth of a network). In the model-free RL setting, we may try to achieve deeper exploration through, for example, parameter space noise over episodes \citep{plappert2017parameter} or through propagation of value uncertainty estimates \citep{osband2016deep,moerland2017efficient}. However, deep exploration is natural to model-based RL, since the planning cycle can perform a deeper lookahead, to which we can then commit in the real environment \citep{lowrey2018plan,sekar2020planning}. Note that for model-based exploration there is one caveat: when we plan for a deep sequence, but then only execute the first action of the sequence and replan (a receding-horizon), we still have the risk of jittering behaviour.  

\item {\it Task-conflated versus task-separated exploration back-ups}: Once we identify an interesting new state (e.g., because it is novel), we want to back-up this information to potentially return there in a next episode. Therefore, back-ups are a crucial element of the exploration cycle. Many intrinsic motivation approaches use intrinsic rewards \citep{chentanez2005intrinsically} (e.g., for novelty), and simply add these as a bonus to the extrinsic reward. The exploration potential of a state is then propagated inside the global value/policy function, together with information about the extrinsic reward. We will call this {\it task-conflated propagation}, since exploration information (intrinsic rewards) is merged with information about the true task (extrinsic rewards). A potential downside of this approach is that exploration information modifies the global solution, and, after an intrinsic reward has worn out, it may take time before its effect on the value function has faded out.

As an alternative, we may also use {\it task-separated} exploration back-ups. In this case, the global solution (value or policy function) is explicitly separated from the exploration information, like the way to get back to a particular interesting region. For example, \citet{shyam2019model} propose to train separate value functions for the intrinsic and extrinsic rewards. We can also use a goal-conditioned policy or value function, which automatically separates information for each potential goal state. In general, task-separated exploration back-ups come at additional computational (and memory) cost, but they do allow for better separation of exploration information and the true extrinsic task. 

\item {\it Parametric versus non-parametric (deep) exploration back-ups}: Similar to the depth of exploration in the forward sense, the depth of the back-up also plays a crucial role for exploration. Imagine our agent just discovered an interesting novel state, which we would like to visit again in a next episode. However, we use a one-step back-up (which only propagates information about this state one step back), and we store this information in a deep policy or value network, to which we make small updates. A a consequence of these choices, information may not propagate far enough (only one step), and/or the change to the global value or policy function may not be large enough to change the behaviour of the agent at the start states. The effect is that the agent has actually found an interesting new region, but due to its type of back-up is not able to directly visit this region again. \citet{ecoffet2019go} call this the `detachment' problem, since the information does not propagate far enough, and the agent therefore detaches from it in its initial states. 

A potential solution to this problem is the use of deeper back-ups, especially in combination with {\it semi-parametric} or {\it non-parametric} representation for the exploratory information. In the context of reinforcement learning, non-parametric representations are better known as {\it episodic memory} \citep{blundell2016model}. In episodic memory, we store the exact action sequence towards a particular state, based on a non-parametric overwrite of information in a table. Note that this approach can also be extended to the semi-parametric setting, where we train a neural network to read and write to this table \citep{graves2014neural,pritzel2017neural}. The benefit of episodic memory is a fast change of information, which can help the agent to directly get back to an interesting region (by simply copying the actions that previously brought it there). Indeed, episodic memory is known to play an important role in human and animal learning as well \citep{gershman2017reinforcement}. Go-Explore \citep{ecoffet2019go} implicitly uses this idea, but directly resets the agent to a previously seen state (without actually replaying the action sequence from the start). Note that there is a variety of other research into episodic memory for reinforcement learning \citep{blundell2016model,pritzel2017neural,lin2018episodic,loynd2018now,ramani2019short,fortunato2019generalization,hu2021generalizable}.
\end{itemize}

\begin{figure}[t]
  \centering
      \includegraphics[width = 0.9\textwidth]{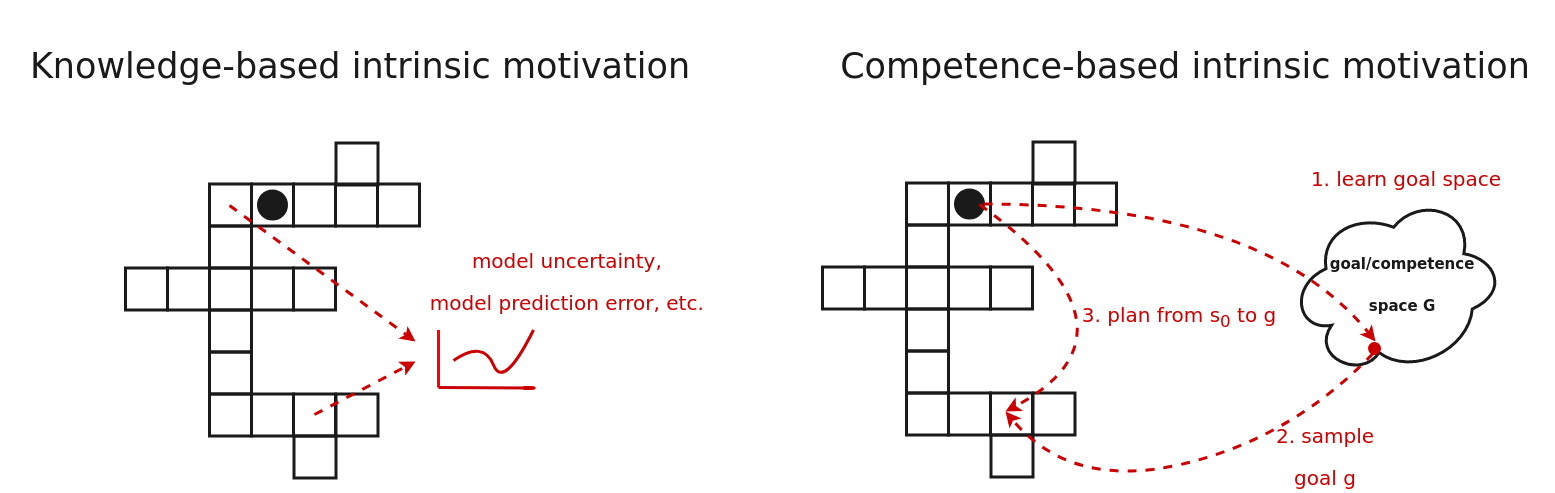}
  \caption{Knowledge-based versus competence-based intrinsic motivation. Solid circle identifies the current agent position. {\bfseries Left}: In knowledge-based intrinsic motivation, every state (the arrows show two examples) in the domain gets associated with an intrinsic reward based on local characteristics, like visitation frequency, uncertainty of the model, prediction error of the model, etc. {\bfseries Right}: In competence-based intrinsic motivation, we learn some form of a goal-space that captures (and compresses) the directions of variation in the domain. We then sample a new goal, for example at the edge of our current knowledge base, and explicitly try to reach it, re-using the way we previously got close to that state.}
    \label{fig_intrinsic_motivation}
\end{figure}

\noindent With our understanding of the above concepts, we are now ready to discuss model-based exploration. We will focus on the intrinsic motivation (IM) literature (i.e., state-based exploration, the bottom row of Table \ref{table_exploration_methods}) \citep{chentanez2005intrinsically}. This field is traditionally split up in two sub-fields \citep{oudeyer2008can}: {\it knowledge-based} and {\it competence-based} intrinsic motivation (Fig. \ref{fig_intrinsic_motivation}), which differ in the way they define the exploration potential of a certain state. 

\paragraph{Knowledge-based intrinsic motivation} 
Knowledge-based intrinsic motivation prioritizes states for exploration when they {\it provide new information about the MDP}. Most approaches in this category specific a specific intrinsic reward, which is then propagated together with the extrinsic reward (task-conflated exploration back-ups). Writing $r^i(s)$ or $r^i(s,a,s')$ for specific intrinsic reward of a certain state or transition, these methods use a total reward that combines the intrinsic and extrinsic part:

\begin{equation}
r_t(s,a,s') = r^e(s,a,s') + \eta \cdot r^i(s,a,s'), \label{eq_kbim}
\end{equation}

where $r^e$ denotes the external reward, and $\eta \in \mathbb{R}$ is a hyperparameter that controls the relative strength of the intrinsic motivation. 

Most knowledge-based IM literature focuses on different ways to specify $r^i$. By far the largest category uses the concept of {\it novelty} \citep{hester2012intrinsically,bellemare2016unifying,sequeira2014learning}. For example, the Bayesian Exploration Bonus (BEB) \citep{kolter2009near} uses 
\begin{equation}
r^i(s,a,s') \propto 1/(1+n(s,a)),
\end{equation}

where $n(s,a)$ denotes the number of visits to state-action pair $(s,a)$. Novelty ideas were studied in high-dimensional problems as well, using the concept of pseudo-counts, which closely mimick density estimates \citep{bellemare2016unifying,ostrovski2017count}. 

There are various other ways to specify the intrinsic reward signal. Long before the term knowledge-based IM became established, \citet{sutton1990integrated} already included an intrinsic reward for {\it recency}: 

\begin{equation}
r^i(s,a,s') = \sqrt{l(s,a)},
\end{equation} 

where $l(s,a)$ denotes the number of timesteps since the last trial at $(s,a)$. More recent examples of intrinsic rewards include {\it model prediction error} \citep{stadie2015incentivizing,pathak2017curiosity}, {\it surprise} \citep{achiam2017surprise}, {\it information gain} \citep{houthooft2016vime}, and {\it feature control} (the ability to change elements of our state over time) \citep{dilokthanakul2019feature}. Note that intrinsic rewards for recency and model prediction error may help overcome non-stationarity (Sec. \ref{sec_non_stationarity}) as well \citep{lopes2012exploration}. Multiple intrinsic rewards can also be combined, like a combination of novelty and model uncertainty \citep{hester2012intrinsically}. Note that many of these intrinsic motivation ideas can be related to emotion theory, which was surveyed for RL agents by \citet{moerland2018emotion}. 

Many of the above knowledge-based IM methods are implemented in a one-phase way, i.e., the intrinsic reward is computed when encountered, but there is not explicit planning towards it. We can of course also combine knowledge-based IM with two-phase exploration \citep{sekar2020planning}, i.e. `plan to explore'. As mentioned before, nearly all knowledge-based IM approaches use task-conflated propagation, while \citet{shyam2019model} do learn separate value functions for the intrinsic and extrinsic rewards. Note that novelty is also an important concept in theoretical work on the sample complexity of exploration \citep{kakade2003sample,brafman2002r}, which we further discuss in Sec. \ref{sec_theory}.

\paragraph{Competence-based intrinsic motivation}
Competence-based intrinsic motivation builds on the same curiosity principles as knowledge-based IM. However, competence-based IM selects new exploration targets based on {\it learning progress}, which focuses on the agent's competence to achieve something, rather than knowledge about the MDP (e.g., we may have visited a state often, which would make it uninteresting for knowledge-based IM, but if we are still getting better/faster at actually reaching the state, i.e., we still make learning progress, then the state does remain interesting for competence-based IM). In competence-based IM the intention is usually to generate an automatic {\it curriculum} of tasks, guided by learning progress \citep{bengio2009curriculum}. 

A popular formulation of compentence-based IM methods are Intrinsically Motivated Goal Exploration Processes (IMGEP) \citep{baranes2009r}, which consist of three steps: 1) learn a goal space, 2) sample a goal, and 3) plan/get towards the goal. Goal space learning was already discussed in Sec. \ref{sec_state_abstraction} and \ref{sec_action_abstraction}. The general aim is to learn a representation that captures the salient directions of variation in a task. For competence-based IM, it may be useful to learn a {\it disentangled} representation, where each controllable object is captured by a separate dimension in the representation. Then, we can create a better curriculum by sampling new subgoals that alter only one controllable object at a time \citep{laversanne2018curiosity}. 

The second step, goal space sampling, is a crucial part of competence-based IM, since we want to select a goal that has high potential for {\it learning progress} \citep{oudeyer2007intrinsic,baranes2013active}. One approach is to track a set of goals, and reselect those goals for which the achieved return has shown positive change recently \citep{matiisen2017teacher,laversanne2018curiosity}. As an alternative, we may also fit a generative model to sample new goals from, which may for example be trained on all previous goals \citep{pere2018unsupervised} or on a subset of goals of intermediate difficulty \citep{florensa2018automatic}. Note that the concept of learning progress has also appeared in knowledge-based IM literature \citep{schmidhuber1991curious}. 

In the third step, we actually attempt to reach the sampled goal. The key idea is that we should already know how to get close to the new goal, since we sampled it close to a previously reached state. Goal-conditioned value functions (discussed in Sec. \ref{sec_action_abstraction}) can be one way to achieve this, but we may also attempt to learn a mapping from current state and goal to policy parameters \citep{laversanne2018curiosity}. Episodic memory methods could also be applied here. 

\vspace{0.3cm}

In short, model-based exploration is an active research topic, which has already made important contributions to exploration research. An important next step would be to show that these methods can also outperform model-free RL in large applications. Another important aspect, which we have not discussed in this section, is the potential benefit of hierarchical RL for exploration. We already covered the challenge of learning good hierarchical actions in Sec. \ref{sec_action_abstraction}. However, once good abstract actions are available, they will likely be a crucial component of (model-based) exploration as well, due to a reducing of the lookahead and propagation depth.

\subsection{Optimality} \label{sec_stability}
Another benefit of model-based RL, in the context of a known model, seems better asymptotic performance. For model-based RL with a learned model, the common knowledge is that we may improve data efficiency, but lose asymptotic performance in the long run.  However, recent attempts of model-based RL with a known model, like AlphaGo Zero \citep{silver2017mastering} and Guided Policy Search \citep{levine2013guided}, manage to outperform model-free attempts on long-run empirical performance. MuZero \citep{schrittwieser2019mastering}, which uses a (value-equivalent) learned model, further outperforms the results of AlphaGo Zero. This suggests that with a perfect (or good) model, model-based RL may actually lead to better (empirical) asymptotic performance. 

A possible explanation for the mutual benefit of planning and learning originates from the type of representation they use. The atomic (tabular) representation of planning does not scale to large problems, since the table would grow too large. The global approximation of learning provides the necessary generalization, but will inevitably make local approximation errors. However, when we add local planning to learning, the local representation may help to locally smooth out the errors in the function approximation, by looking ahead to states with more clearly discriminable value predictions. These local representations are often tabular/exact, and can thereby give better local separation. For example, in Chess the learned value prediction for the current state of the board might be off, but through explicit lookahead we may find states that are a clear win or loss in a few steps. As such, local planning may help learning algorithms to locally smooth out the errors in its approximation, leading to better asymptotic performance.

There is some initial work that supports these ideas. \citet{silver2008sample} already described the use of {\it transient} and {\it permanent} memory, where the transient memory is the local plan that fine-tunes the value estimates. Both \citet{moerland2020think} and \citet{langlois2019benchmarking} recently studied the trade-off between planning and learning (already mentioned in Sec. \ref{sec_tradeoff}), finding that optimal performance requires an intermediate planning budget per real step, and not a high budget (exhaustive search), or no planning budget per timestep at all (model-free RL). Since model-free RL is notoriously unstable in the context of function approximation \citep{henderson2018deep}, we may hypothesize that the combination of global function approximation (learning) and local atomic/tabular representation (planning) helps stabilize learning and achieve better asymptotic performance (see \citet{hamrick2020combining} as well). 

To conclude, we note that this combination of local planning and global approximation also exists in humans. In cognitive science, this idea is known as dual process theory \citep{evans1984heuristic}, which was more recently popularized as `thinking fast and slow' \citep{kahneman2011thinking}. \citet{anthony2017thinking} connect planning-learning integration to these ideas, suggesting that global policy or value functions are like `thinking fast', while local planning relates to explicit reasoning and `thinking slow'. 

\subsection{Transfer} \label{sec_transfer}
In {\it transfer learning} \citep{taylor2009transfer,lazaric2012transfer} we re-use information from a source task to speed-up learning on a new task. The source and target tasks should neither be the same, as then transfer is trivial, nor completely unrelated, as then there is no information to transfer. \citet{konidaris2006framework} covers a framework for transfer, specifying three types: i) transfer of a dynamics model, ii) transfer of skills or sub-routines, and iii) transfer of `knowledge', like shaping rewards and representations. For this model-based RL survey we only discuss the first category, transfer of a dynamics model. There are largely two scenarios: i) similar dynamics function but different reward function, for example a new level in a video game, and ii) slightly changed transition dynamics, for example transfer from simulation to real-world tasks. We discuss examples in both categories. 

\paragraph{Same dynamics with different reward} The first description of model transfer with a changed reward function is by \citet{atkeson1997comparison}. The authors change the reward function in a Pendulum swing-up task after 100 trials, and show that the model-based approach is able to adapt much faster, requiring less data from the real environment. Later on, the problem (different reward function with stationary dynamics) became better known as {\it multi-objective} reinforcement learning (MORL) \citep{roijers2017multi,roijers2013survey}. A multi-objective MDP has a single dynamics function but multiple reward functions. These rewards can be combined in different ways, each of which lead to a new task specification. There are many model-free approaches for the MORL setting \citep{roijers2013survey}, with model-based examples given by \citet{wiering2014model}, \citet{yamaguchi2019model}. Other examples of model-based transfer to different reward functions (goals) are provided by \citet{sharma2019dynamics} and \citet{sekar2020planning}. 

Another approach designed for changing reward functions is the successor representation \citep{dayan1993improving,barreto2017successor}. Successor representations summarize the model in the form of future state occupancy statistics. It thereby falls somewhere in between model-free and model-based methods \citep{momennejad2017successor}, since these methods can partially adapt to a different reward function, but it does not fully compute new occupancy statistics like a full model-based method would. 

\paragraph{Different dynamics} In the second category we find transfer to a task with slightly different dynamics. Conceptually, \citet{konidaris2007building} propose to disentangle the state into an agent space (which can directly transfer) and a problem space (which defines the new task). However, disentanglement of agent and problem space is still hard without prior knowledge. 

One way to achieve good transfer is by learning representations that generalize well. The object-oriented and physics-based approaches, already introduced in Sec. \ref{sec_state_abstraction}, have shown success in achieving this. For example, Schema Networks \citep{kansky2017schema} learn object interactions in Atari games, and manage to generalize well to several variations of Atari Breakout, like adding a new wall or slightly changing the dynamics (while still complying with the overall physics rules). 

Simulation-to-real transfer is popular in robotics, but most researchers transfer a policy or value function \citep{tobin2017domain}. Example approaches that do transfer a dynamics model to the real world are \citet{christiano2016transfer} and \citet{nagabandi2018learning}. Several researchers also take a zoomed out view, where they attempt to learn a distribution over the task space, better known as {\it multi-task learning} \citep{caruana1997multitask}. Then, when a new task comes in, we may quickly identify in which cluster of known tasks (dynamics models) it belongs \citep{wilson2007multi}. Another approach is to learn a global neural network initialization that can quickly adapt to new tasks sampled from the task space \citep{clavera2018model}, which implicitly transfers knowledge about the dynamics of related tasks. 

\vspace{0.3cm}

\noindent In short, transfer is one of the main benefits of model-based RL. \citet{van2020loca} even propose a metric, the Local Change Adaptation (LoCA) regret, to compare model-based RL algorithms based on their ability to learn on new, slightly altered tasks. An overview of transfer methods for deep reinforcement learning in general is provided by \citet{zhu2020transfer}. 

\subsection{Safety} \label{sec_safety}
Safety is an important issue, especially when learning on real-world systems \citep{amodei2016concrete}. For example, with random exploration it is easy to break a robot before any learning has taken place. \citet{berkenkamp2017safe} studies a model-based safe exploration approach based on the notion of asymptotic stability. Given a `safe region' of the current policy, we want to explore while ensuring that we can always get back to the safe region. As an alternative, \citet{aswani2013provably} keep two models: the first one is used to decide on an exploration policy, while the second model has uncertainty bounds and is used for verification of the safety of the proposed policy. \citet{ostafew2016robust} ensure constraints by propagating uncertainty information in a Gaussian Process model. Safety is a vital aspect of real-world learning, and it may well become an important motivation for model-based RL in forthcoming years.

\subsection{Explainability} \label{sec_explainability}
Explainable artificial intelligence (XAI) has received much attention in the AI community in recent years. Explainable reinforcement learning (XRL) was studied by \citet{van2018contrastive}, who generated explanations from planned traces. The authors also study contrastive explanations, where the user can ask the agent why it did not follow another policy. There is also work on RL agent transparency based on emotion elicitation during learning \citep{moerland2018emotion}, which largely builds on model-based methods. Finally, \citet{shu2017hierarchical} study language grounding in reinforcement learning, which is an important step to explainability as well. Explainability is now widely regarded as a crucial prerequisite for AI to enter society. Model-based RL may be an important element of explainability, since it allows the agent to communicate not only its goals, but also the way it intends to achieve them. 

\subsection{Disbenefits}
Model-based RL has disbenefits as well. First, model-based RL typically requires additional computation, both for training the model, and for the planning operations themselves. Second, model-based RL methods with a learned model can be unstable due to uncertainty and approximation errors in the model. Therefore, although these approaches can be more data efficient, they also tend to have lower asymptotic performance. We already extensively discussed how to deal with model uncertainty. Third, model-based RL methods require additional memory, for example to store the model. However, with function approximation this is typically not a large limitation. Finally, model-based RL algorithms typically have more tunable hyperparameters than model-free algorithms, including hyperparameters to estimate uncertainty, and hyperparameters to balance planning and real data collection. Most of these disbenefits are inevitable, and we are essentially trading extra computation, memory and potential instability (for a learned model) against better data efficiency, targeted exploration, transfer, safety and explainability. 

\section{Theory of Model-based Reinforcement Learning} \label{sec_theory}
There is also a large body of literature on the theoretical convergence properties of model-based reinforcement learning. Although the primary focus of this survey was on the practical/empirical aspects of model-based RL, we will shortly highlight some main theoretical results. Classic convergence results in dynamic programming are for example available for policy iteration \citep{puterman2014markov}, approximate policy iteration \citep{kakade2002approximately,munos2003error}, and real-time dynamic programming (RTDP) \citep{barto1995learning}. \citet{efroni2018beyond} show multi-step policy iteration also converges, as does multi-step and approximate RTDP \citep{efroni2019multi}.

Much theoretical work tries to quantify the rate at which algorithms converge, which we can largely split up in sample complexity bounds (PAC bounds) and regret bounds. Sample complexity is typically assessed through the Probably Approximately Correct (PAC) framework. Here, we try to bound the number of timesteps an algorithm may select an actions whose value is not near-optimal \citep{strehl2006pac}. An algorithm is PAC if this number is bounded by a function polynomial in the problem characteristics, like the MDP horizon $H$ and the dimensionality of the state ($|\mathcal{S}|$) and action space ($|\mathcal{A}|$). There are a variety of papers that provide PAC bounds for MDP algorithms \citep{kakade2003sample,strehl2006pac,dann2015sample,asmuth2009bayesian,szita2010model}.

An alternative approach is to bound the {\it regret} during the learning process. The regret measures the average total loss of reward of the learned policy compared to the optimal policy. The regret at timestep $T$ is therefore defined as

\begin{equation}
\text{Regret}(T) = \mathbb{E}_{\pi^\star} [ \sum_{t=1}^T r_t ] -  \mathbb{E}_{\pi} [ \sum_{t=1}^T r_t ],
\end{equation}

where $\pi^\star$ denotes the optimal policy, and $\pi$ denotes the (changing) policy of our learning algorithm. While PAC bounds the {\it total number} of sub-optimal actions a learned policy will take, regret bounds limit the {\it total size} of the mistakes. The {\it lower} bound of the above regret is known to be $\Omega(\sqrt{H |\mathcal{S}| |\mathcal{A}| T})$ \citep{jaksch2010near,osband2016lower}. Table \ref{table_regret_bounds} lists several model-based RL algorithms with proven {\it upper} regret bounds. UCRL2 \citep{jaksch2010near} obtains a regret bound of $\tilde{O}(H |\mathcal{S}| \sqrt{|\mathcal{A}| T})$, which \citet{agrawal2017posterior} improve to $\tilde{O}(H \sqrt{|\mathcal{S}| |\mathcal{A}| T})$ for large $T$. UCBVI \citep{azar2017minimax} and vUCQ \citep{kakade2018variance} further improve these results, with UCBVI achieving $\tilde{O}( \sqrt{H |\mathcal{S}| |\mathcal{A}| T} + \sqrt{H^2T})$. While these algorithms provide {\it worst-case} regret bounds, EULER \citep{zanette2019tighter} actually matches the theoretical lower bound under additional assumptions on the variance of the optimal value function. 

\begin{table}[t]
\centering
\caption{Minimax regret bounds for different model-based reinforcement learning algorithms. $|\mathcal{S}|$ and $|\mathcal{A}|$ denote the size of the state and action space, respectively. $H$ denotes the MDP horizon, and $T$ is the total number of samples of the algorithm. EULER uses additional assumptions on the variance of the optimal value function.}
\label{table_regret_bounds}
\begin{tabular}{ p{7.0cm} P{5.0cm}}
\toprule
\bf Algorithm  & \bf Regret \\
   \hline

UCRL2 \citep{jaksch2010near} & $\tilde{O}(H |\mathcal{S}| \sqrt{|\mathcal{A}| T})$ \\

\citep{agrawal2017posterior} (for large $T$) & $\tilde{O}(H  \sqrt{|\mathcal{S}||\mathcal{A}| T})$ \\

UCBVI \citep{azar2017minimax} & $\tilde{O}( \sqrt{H |\mathcal{S}| |\mathcal{A}| T} + H \sqrt{T})$ \\

EULER \citep{zanette2019tighter} & $\tilde{O}( \sqrt{H |\mathcal{S}| |\mathcal{A}| T})$ \\

\hline
{\it Model-free} (Q-learning) \citep{jin2018q}  & $\tilde{O}( \sqrt{H^3 |\mathcal{S}| |\mathcal{A}| T})$ \\
{\it Lower bound} \citep{jaksch2010near} &  $\Omega(\sqrt{H |\mathcal{S}| |\mathcal{A}| T})$ \\

\bottomrule
\end{tabular}
\end{table}

All previously discussed algorithms use a variant of {\it optimism in the face of uncertainty} \citep{lai1985asymptotically} in their algorithms, usually through upper confidence bounds on either estimated dynamics models or value functions. There is an alternative Bayesian approach known as {\it posterior sampling reinforcement learning} (PSRL) \citep{osband2013more}, which instead estimates a Bayesian posterior and generally uses Thompson sampling \citep{thompson1933likelihood}. Algorithms in this category \citep{osband2013more,gopalan2015thompson,osband2017posterior} do use a Bayesian formulation of regret, which is less strict than the worst-case (minimax) regret. However, some Bayesian approaches  also come with frequentist worst-case regret bounds \citep{agrawal2017posterior}.

Note that there are also regret bounds for model-free RL algorithms \citep{jin2018q} (Table \ref{table_regret_bounds}), but these fall outside the scope of this survey. The main overall story is that model-based RL allows for better PAC/regret bounds than model-free RL, but also suffers from worse (computational) time and space complexity (since we need to estimate and store the transition function). Moreover, all previously discussed methods assume full planning, which also adds to the computational burden. However, \citet{efroni2019tight} interestingly show that one-step greedy planning, as for example used in Dyna \citep{sutton1990integrated}, can actually match the regret bounds of UCRL2 and EULER, while reducing its time and space complexity.  

\section{Related Work} \label{sec_relatedwork}
While model-based RL has been successful and received much attention \citep{silver2017mastering,levine2013guided,deisenroth2011pilco}, a survey of the field currently lacks in literature. \citet{hester2012learning} gives a book-chapter presentation of model-based RL methods, but their work does not provide a full overview, nor does it incorporate the vast recent literature on neural network approximation in model-based reinforcement learning.

\citet{moerland2020frap} present a framework for reinforcement learning and planning that disentangles their common underlying dimensions, but does not focus on their integration. In some sense, \citet{moerland2020frap} look `inside' each planning or reinforcement learning cycle, strapping their shared algorithmic space down into its underlying dimensions. Instead, our work looks `over' the planning cycle, focusing on how we may integrate planning, learning and acting to provide mutual benefit. 

\citet{hamrick2019analogues} presents a recent coverage of mental simulation (planning) in deep learning. While technically a model-based RL survey, the focus of \citet{hamrick2019analogues} lies with the relation of these approaches to cognitive science. Our survey is more extensive on the model learning and integration side, presenting a broader categorization and more literature. Nevertheless, the survey by \citet{hamrick2019analogues} is an interesting companion to the present work, for deeper insight from the cognitive science perspective. \citet{plaat2020model} also provide a recent description of model-based RL in high-dimensional state spaces, and puts additional emphasis on implicit and end-to-end model-based RL (see Sec. \ref{sec_implicit_mbrl} as well). 

Finally, several authors \citep{nguyen2011model,polydoros2017survey,sigaud2011line} have specifically surveyed structured model estimation in robotics and control tasks. In these cases, the models are structured according to the known laws of physics, and we want to estimate a number of free parameters in these models from data. This is conceptually similar to Sec. \ref{sec_model_learning}, but our work discusses the broader supervised learning literature, when applicable to dynamics model learning. Thereby, the methods we discuss do not need any prior physics knowledge, and can deal with much larger problems. Moreover, we also include discussion of a variety of other model learning challenges, like state and temporal abstraction.

\section{Discussion} \label{sec_discussion}
This chapter surveyed the full spectrum of model-based RL, including model learning, planning-learning integration, and the benefits of model-based RL. To further advance the field, we need to discuss two main topics: benchmarking, and future research directions. 

\paragraph{Benchmarking}
Benchmarking is crucial to the advancement of a field. For example, major breakthroughs in the computer vision community followed the yearly ImageNet competition \citep{krizhevsky2012imagenet}. We should aim for a similar benchmarking approach in RL, and in model-based RL in particular. 

A first aspect of benchmarking is proper assessment of problem difficulty. Classic measures involve the breadth and depth of the full search tree, or the dimensionality of the state and action spaces. While state dimensionality was for long the major challenge, breakthroughs in deep RL are now partially overcoming this problem. Therefore, it is important that we start to realize that state and action space dimensionality are not the only relevant measures of problem difficulty. For example, sparse reward tasks can be challenging for exploration, even in low dimensions. \citet{osband2019behaviour} recently proposed a benchmarking suite that disentangles the ability of an algorithm to deal with different types of challenges. 

A second part of benchmarking is actually running and comparing algorithms. Although many benchmarking environments for RL have been published in recent years \citep{bellemare2013arcade,brockman2016openai}, and benchmarking of model-free RL has become quite popular, there is relatively little work on benchmarking model-based RL algorithms. \citet{langlois2019benchmarking} recently made an important first step in this direction by benchmarking several model-based RL algorithms, and the field would profit from more efforts like these. 

For reporting results, an important remaining challenge for the entire RL community is standardization of learning curves and results. The horizontal axis of a learning curve would ideally show the number of unique flops (computational complexity) or the number of real world or model samples. However, many papers report `training time in hours/days' on the horizontal axis, which is of course heavily hardware dependent. Other papers report `episodes' on the horizontal axis, while a model-based RL algorithm uses much more samples than a model-free algorithm per episode. When comparing algorithms, we should always aim to keep either the total computational budget or the total sample budget equal.

\paragraph{Future work}
There is a plethora of future work directions in the intersection of planning and learning. We will mention a few research areas, which already received much attention, but have the potential to generate breakthroughs in the field. 

\begin{itemize}
\item Asymptotic performance: Model-based RL with a learned model tends to have better sample complexity, but inferior asymptotic performance, compared to model-free RL. This is an important limitation. AlphaGo Zero recently illustrated that model-based RL with a known model should be able to surpass model-free RL performance. However, in the context of a learned model, a major challenge is to achieve the same optimal asymptotic performance as model free RL, which probably requires better ways of estimating and dealing with model uncertainty. 

\item Hierarchy: A central challenge, which has already received much attention, is temporal abstraction (hierarchical RL). We still lack consistent methods to identify useful sub-routines, which compress, respect reward relevancy, identify bottleneck states and/or focus on interaction with objects and salient domain aspects. The availability of good temporal abstraction can strongly reduce the depth of a tree search, and is likely a key aspect of model-based learning. 

\item Exploration \& Competence-based intrinsic motivation: A promising direction within exploration research could be competence-based intrinsic motivation \citep{oudeyer2007intrinsic}, which has received less attention than its brother knowledge-based intrinsic motivation (see Sec. \ref{sec_exploration}). By sampling goals close to the border of our currently known set, we generate an automated curriculum, which may make exploration more structured and targeted. 

\item Transfer: We believe model-based RL could also put more emphasis on the transfer setting, especially when it comes to evaluating data efficiency. It can be hard to squeeze out all information on a single, completely new task. Humans mostly use forward planning on reasonably certain models that generalize well from previous tasks. Shifting RL and machine learning from single task optimization to more general artificial intelligence, operating on a variety of tasks, is an important challenge, in which model-based RL may definitely play an important role. 

\item Balancing: Another important future question in model-based RL is balancing planning, learning and real data collection. These trade-offs are typically tuned as hyperparameters, which seem to be crucial for algorithm performance \citep{langlois2019benchmarking,moerland2020think}. Humans naturally decide when to start planning, and for how long \citep{kahneman2011thinking}. Likely, the trade-off between planning and learning should be a function of the collected data, instead of a fixed hyperparameter.

\item Prioritized sweeping: Prioritized sweeping has been successful in tabular settings, when the model is trivial to revert. As mentioned throughout the survey, it has also been applied to high-dimensional approximate settings, but this creates a much larger challenge. Nevertheless, exploration in the forward direction may actually be just as important as propagation in the backwards direction, and prioritized sweeping in high-dimensional problems is definitely a topic that deserves attention. 

\item Optimization: Finally, note that RL is effectively an optimization problem. While this survey has focused on the structural aspects of this challenge (what models to specify, how to algorithmically combine them, etc.), we also observe much progress in combining optimization methods, like gradient descent, evolutionary algorithms, automatic hyperparameter optimization, etc. Such research may have an equally big impact on progress in MDP optimization and sequential decision making.
\end{itemize}

\section{Summary} \label{sec_summary}

This concludes our survey of model-based reinforcement learning. We will briefly summarize the key points:

\begin{itemize}
\item Nomenclature in model-based RL is somewhat vague. We define model-based RL as `any MDP approach that uses i) a model (known or learned) and ii) learning to approximate a global value or policy function'. We distinguish three categories of planning-learning integration: `model-based RL with a learned model', `model-based RL with a known model', and `planning over a learned model' (Table \ref{tab_model_based_boundaries2}).
\item Model-based reinforcement learning may first require approximation of the dynamics model. Key challenges of model learning include dealing with: environment stochasticity, uncertainty due to limited data, partial observability, non-stationarity, multi-step prediction, and representation learning methods for state and temporal abstraction (Sec. \ref{sec_model_learning}). 
\item Integration of planning and learning involves a few key aspects: i) where to start planning, ii) how much budget to allocate to planning and acting, iii) how to plan, and iv) how to integrate the plan in the overall learning and acting loop. Planning-learning methods widely vary in their approach to these questions (Sec. \ref{sec_model_using}). 
\item Explicit model-based RL manually designs model learning, planning algorithms and the integration of these. In contrast, implicit model-based RL optimizes elements of this process, or the entire model-based RL computation, against the ability to predict an outer objective, like a value or optimal action (Sec. \ref{sec_implicit_mbrl}). 
\item Model-based RL can have various benefits, including aspects like data efficiency, targeted exploration, transfer, safety and explainability (Sec. \ref{sec_benefits}). Recent evidence indicates that the combination of planning and learning may also provide more stable learning, possibly due to the mutual benefit of global function approximation and local tabular representation.   
\end{itemize}

\vspace{0.4cm}

\noindent In short, both planning and learning are large research fields in MDP optimization that depart from a crucially different assumption: the type of access to the environment. Cross-breeding of both fields has been studied for many decades, but a systematic categorization of the approaches and challenges to model learning and planning-learning integration lacked so far. Recent examples of model-based RL with a known model \citep{silver2017mastering,levine2013guided} have shown impressive results, and suggest much potential for future planning-learning integrations. This survey conceptualized the advancements in model-based RL, thereby: 1) providing a common language to discuss model-based RL algorithms, 2) structuring literature for readers that want to catch up on a certain subtopic, for example for readers from either a pure planning or pure RL background, and 3) pointing to future research directions in planning-learning integration. 

\bibliographystyle{apalike}
\bibliography{overview}
\end{document}